\documentclass[journal]{IEEEtran}
\usepackage{amsmath,epsfig,color}
\usepackage{url}
\usepackage{bigstrut,multirow,rotating}
\usepackage{textcomp,booktabs}
\usepackage{tabularx}
\usepackage{floatrow}
\newfloatcommand{capbtabbox}{table}[][\FBwidth]
\usepackage{caption}
\usepackage{colortbl}
\usepackage{xcolor}
\usepackage{multirow}
\usepackage{booktabs}
\usepackage{underscore}
\usepackage[marginal]{footmisc}
\usepackage{subfigure}
\begin{document}
\newcommand{ \bl }[1]{{\color{blue}#1}}
\newcommand{ \blr}[1]{{\color{red}#1}}
\title{Camera Invariant Feature Learning for Generalized Face Anti-spoofing}
\author{Baoliang Chen, Wenhan Yang,~\IEEEmembership{Member,~IEEE}, 
Haoliang Li, 
Shiqi Wang,~\IEEEmembership{Member,~IEEE},       and  Sam Kwong,~\IEEEmembership{Fellow,~IEEE}
\thanks{B. Chen, W. Yang, S. Wang and S. Kwong are with the Department of Computer Science, City University of Hong Kong, Hong Kong (E-mail addresses: blchen6-c@my.cityu.edu.hk; wyang34@cityu.edu.hk; shiqwang@cityu.edu.hk; cssamk@cityu.edu.hk). S. Wang is also with Shenzhen Research Institute, City University of Hong Kong, Shenzhen, China. H. Li is with the Rapid-Rich Object Search Lab, Nanyang Technological University, Singapore (E-mail address: hli016@e.ntu.edu.sg). Corresponding author: Shiqi Wang.}

\thanks{This research was supported in part by the Science, Technology, and Innovation Commission of Shenzhen Municipality under Project JCYJ20180307123934031, National Natural Science Foundation of China under 62022002, in part by the Hong Kong RGC ECS under Grant 21211018, GRF under Grant 11203220.}
}


\maketitle

\begin{abstract}
There has been an increasing consensus in learning based face anti-spoofing that the divergence in terms of camera models is causing a large domain gap in real application scenarios. We describe a framework that eliminates the influence of inherent variance from acquisition cameras at the feature level, leading to the generalized face spoofing detection model that could be highly adaptive to different acquisition devices. In particular, the 
framework is composed of two branches. The first branch aims to learn the camera invariant spoofing features via feature level decomposition in the high frequency domain. Motivated by the fact that the spoofing features exist not only in the high frequency domain, in the second branch the discrimination capability of extracted spoofing features is further boosted from the enhanced image based on the recomposition of the high-frequency and low-frequency information.
Finally, the classification results of the two branches are fused together by a weighting strategy. 
Experiments show that the proposed method can achieve better performance in both intra-dataset and cross-dataset settings, demonstrating the high generalization capability in various application scenarios. 
\end{abstract}

\IEEEpeerreviewmaketitle


\section{Introduction}
\IEEEPARstart{F}{ace} authentication services have been growing exponentially in the past decade, coinciding with the accelerated proliferation of acquisition devices and advances of artificial intelligence. 
Though unexceptionable performance has been achieved, the security issue is a very challenging problem as the system can be easily attacked even by a printed photo or replayed video. To prevent the face recognition system from being vulnerable, face Presentation Attacks Detection (PAD) algorithms have been widely studied to distinguish the spoofing faces from the live one.

Recently, various face presentation attacking methods have been developed to deceive the authentication systems, such as print attack (printing a face on a paper), replay attack (replaying a face video by other devices), and mask attack (wearing a mask). In the literature, numerous methods have been investigated for PAD, and the majority of them rely on computational vision algorithms. In particular, both handcrafted \cite{chingovska2012effectiveness,boulkenafet2016face,komulainen2013context,patel2016secure} and deep learning \cite{yang2014learn,patel2016cross, li2016original,feng2016integration} based features have been developed. For deep learning based methods that rely on training of models based upon labelled data, a large gap between the limited performance and essential requirements have been observed when the training and testing data are from different domains. One typical example is that the training data are acquired with one type of camera and testing data are from another type of camera. To increase the domain adaptation and generalization capability, efforts have been devoted to extracting auxiliary information, including depth~\cite{liu2018learning} and Remote Photoplethysmography (rPPG) signals \cite{liu20163d}. Attempts have also been made to use domain adaptation technology \cite{li2018unsupervised,li2018learning,tzeng2017adversarial} for mitigating the gap between different domains.

\begin{figure}[t]
\begin{minipage}[b]{1\linewidth}
  \centering
  \centerline{\includegraphics[width=1\linewidth]{ 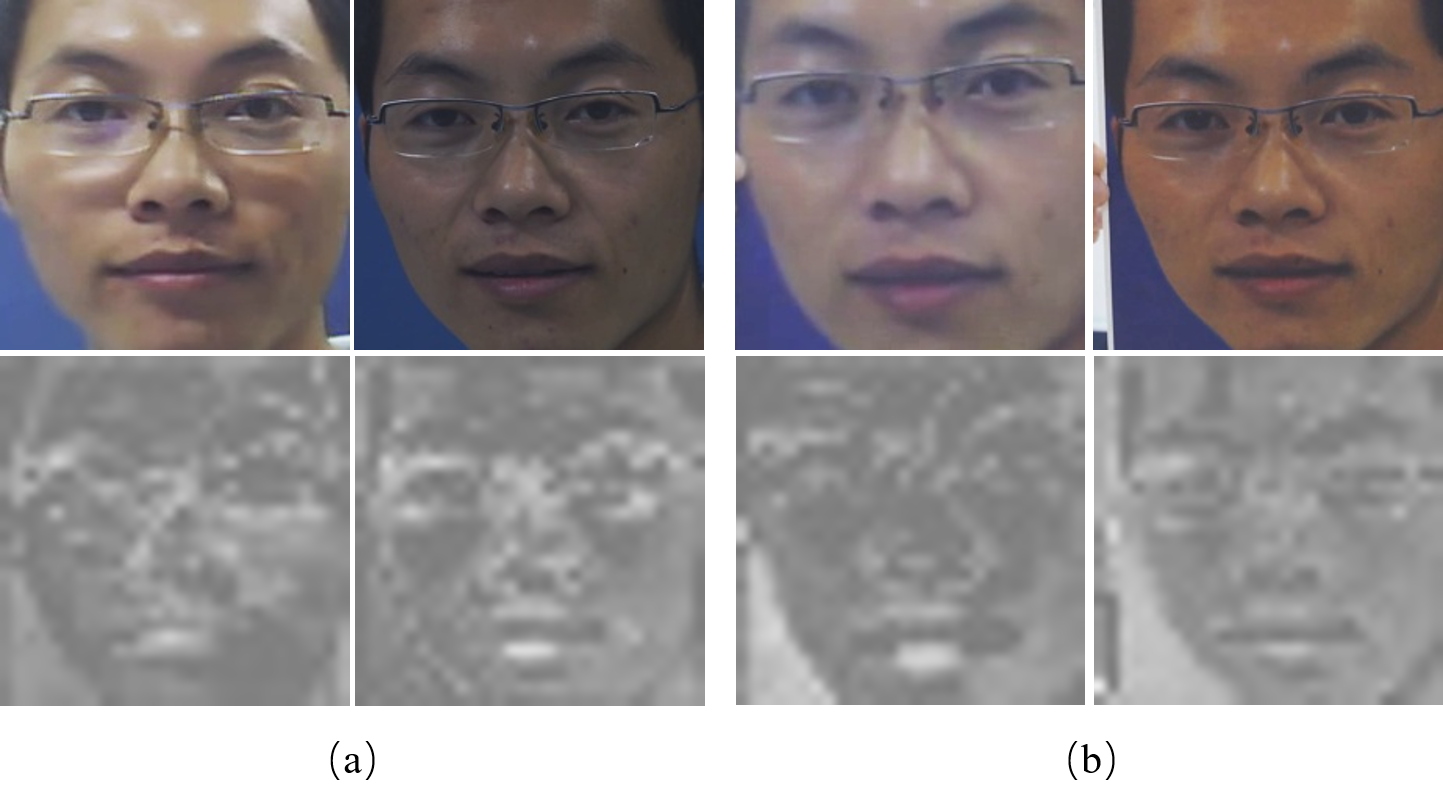}}
\end{minipage}
\caption{Illustrations of live (a) and spoofing (b) faces acquired with two different cameras. The gray-scale maps show corresponding features of the second residual layer of the face images.
Two types of cameras in CASIA-FASD \cite{zhang2012face} database are used for demonstration, including the low quality camera (first and third columns) and high quality camera (second and fourth columns). The features are from the  ResNet-18 \cite{he2016deep} network trained on CASIA-FASD database. }
\label{fig:1}
\end{figure}

\begin{figure}[t]
\begin{minipage}[b]{1\linewidth}
  \centering
  \centerline{\includegraphics[width=0.9\linewidth]{ 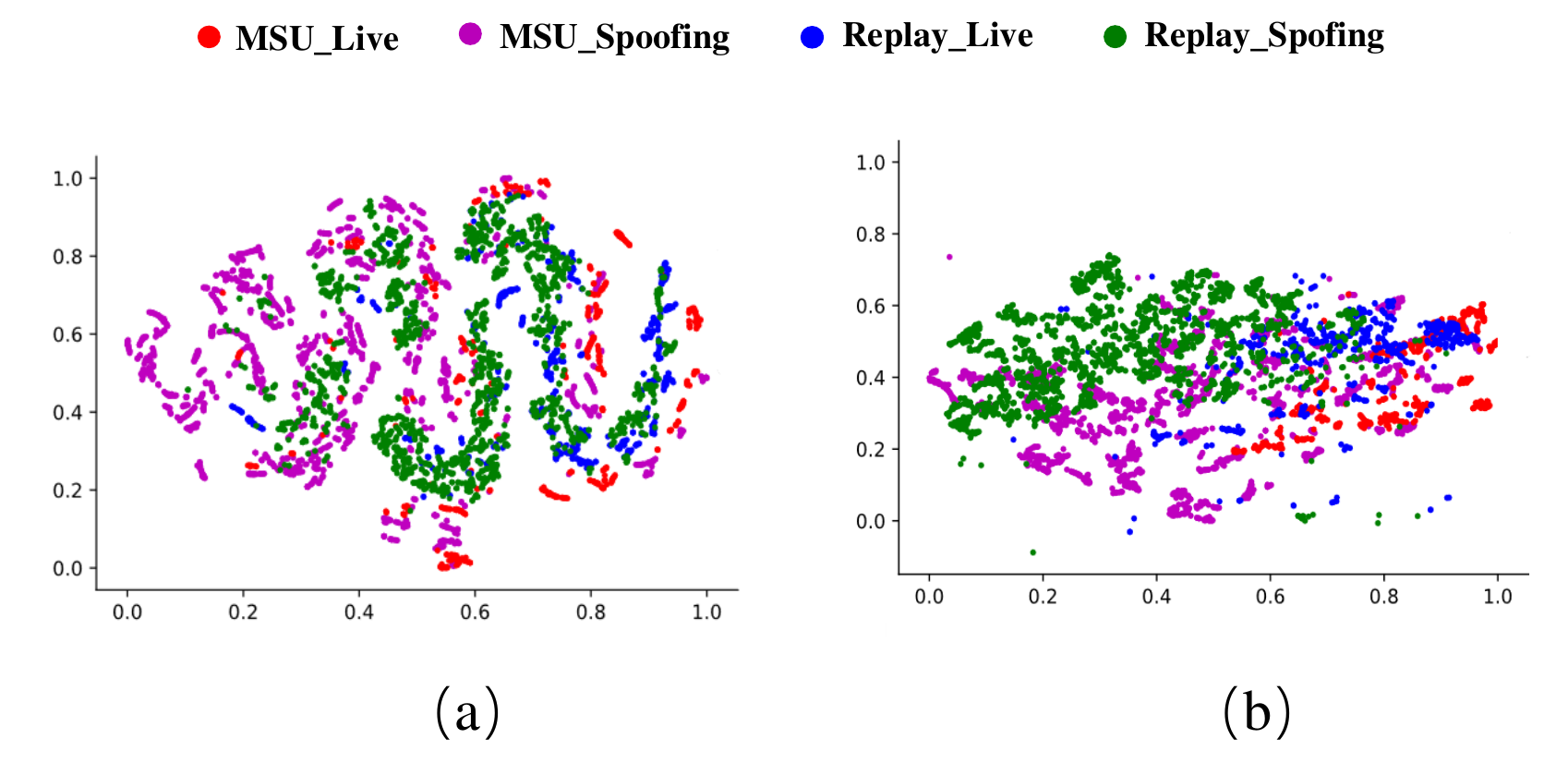}}
\end{minipage}
\caption{T-SNE \cite{maaten2008visualizing} visualization of the second-to-last fully connected layer. (a) ResNet-18; (b) the first branch of our proposed model. The two networks are trained on CASIA-FASD database and tested on two types of cameras in MSU-MFSD \cite{wen2015face} and Replay-Attack \cite{chingovska2012effectiveness} databases.} 
\label{fig:2}
\end{figure}

It has been widely recognized that the camera information is a dominant factor causing the domain gap. One typical example is shown in Fig.~\ref{fig:1}. In particular, although the spoofing face images are generated by an identical attack type (print attack) and identity, the spoofing patterns still dramatically vary according to the cameras. This phenomenon reveals that the camera divergence between training and testing could cause the PAD performance degradation, which is further validated in Fig.~\ref{fig:2}(a). More specifically, it is observed that features from spoofing faces and live faces are lack of discrimination capabilities with large overlapping in-between when training and testing are performed based on cross-camera settings. However, given numerous camera types, it is difficult to collect sufficient training data and specifically train a model for each of them, especially with the surge of emerging cameras. Motivated by this, we aim to propose a novel deep learning based PAD model with high generalization capability. To this end, the camera information should be effectively removed.
The composition of facial features, which has been widely studied in the literature~\cite{chen2012bayesian,jourabloo2018face}, motivates us to automatically learn the camera invariant features. Herein, we propose a feature level decomposition scheme, such that the trained model does not depend on the acquisition device. This allows the model to be widely applied in myriad applications, as the learned model can be well generalized to unseen cameras at large. Extensive experiments have demonstrated that the proposed scheme achieves the state-of-the-art performance and reveals high generalization capability. 
The main contributions of this paper are as follows,
\begin{itemize}
\item We propose a novel framework with two branches to improve the generalization capability of face spoofing.  The first is camera invariant branch, aiming to provide high-frequency domain features with the elimination of camera variance. Considering spoofing features in other frequency domains (\textit{e.g.}, lighting) may be neglected in the first branch, our second branch is feature discrimination augmentation branch which generates features by an enhanced image recomposed from the low and high frequency layers of the original input image.

\item We develop a sophisticated camera variance removal scheme based on the feature level decomposition. The features with the mixture of spoofing and camera information are efficiently decomposed with a pseudo siamese network, in an effort to blindly infer the feature that well reflects the spoofing information while being invariant to different camera types. 

\item The classification results of the two branches are fused for the final decision. Experiments show that our proposed method can achieve high accuracy not only on intra-dataset settings but also on cross-dataset scenarios, demonstrating superior generalization capacity with camera-invariant feature extraction. 
\end{itemize}

The rest of this paper is organized as follows. We first review the related works in Section \uppercase\expandafter{\romannumeral2}. Subsequently, the proposed scheme  is detailed in Section \uppercase\expandafter{\romannumeral3}. The  experimental results are presented in Section \uppercase\expandafter{\romannumeral4}, and finally we conclude this paper in  Section \uppercase\expandafter{\romannumeral5}.

\section{Related Works}
For face anti-spoofing, the intrinsic spoofing features relying on which the binary classification can be performed are expected to be both discriminative and of high generalization capability~\cite{ evans2019handbook}. 
In the literature, a series of features have been studied and developed, including both hand-crafted and deep learning features. 

\subsection{Hand-crafted Features} 
Based on the observation that certain characteristics of texture in spoofing face and live face are different, hand-crafted features are first exploited, including Local Binary Patterns (LBP) \cite{chingovska2012effectiveness}, Local Phase Quantization  (LPQ) \cite{boulkenafet2016face}, Histogram of Gradients (HoG) \cite{komulainen2013context}, Scale-Invariant Feature Transform  (SIFT) \cite{patel2016secure} and Speeded Up Robust Features (SURF) \cite{boulkenafet2016face}. In contrast with the feature extraction performed in spatial domain, in \cite{li2004live}, Li \textit{et al.} utilized the dissimilarity in Fourier spectra by considering that less High Frequency (HF) components exist in spoofing images compared with the live ones. To obtain texture features based on the 3-D plane in videos, the high frequency information in both spatial and temporal domains is exploited in \cite{de2014face}. In \cite{chan2017face}, Chan \textit{et al.}  incorporated flash light for more stable spoofing feature extraction by reducing the influence of environmental factors. Compared with HF texture features, Low Frequency (LF) features have also been utilized together in image quality based methods \cite{galbally2014face,galbally2013image,feng2016integration}. In~\cite{galbally2013image}, with the live face image as the reference, color distortion relevant features are extracted and compared by Mean Squared Error (MSE), Maximum Difference (MD), R-Averaged Maximum Difference (RAMD), \textit{etc}. Regarding no-reference image quality assessment, in \cite{wen2015face}, the concatenated features of specular, blurriness and color distortion are utilized.

Although those handcrafted features are computationally efficient and perform well in intra settings, they may easily fail when there are large variations in terms of attack scenarios~\cite{boulkenafet2018generalization}. To tackle this issue, additional clues such as motion, depth and blood circulation have also been incorporated. Motion clues from eye blinking \cite{pan2008liveness,tirunagari2015detection} and word speaking \cite{bharadwaj2013computationally,siddiqui2016face} can be acquired from multi-frames. Moreover, in \cite{chingovska20132nd}, the pulse generated by facial blood circulation is used since only the live face videos have such traits. In~\cite{bharadwaj2013computationally}, the facial expression clues were firstly enhanced by an Eulerian motion magnification algorithm, then the LBP texture features and Histograms of Oriented Optical Flow (HOOF) motion features were fused together for final classification. The 3D depth information of the captured face \cite{wang2013face}, \cite{wang2017robust} and infrared images \cite{zhang2011face} are effective clues though these solutions rely on additional sensors and could be more expensive to launch.

\subsection{Deep Learning Features} 
For face PAD, deep learning based methods have also been widely studied for obtaining more discriminative features that account for the spoofing patterns. Yang \textit{et al.} \cite{yang2014learn} first proposed to use Convolutional Neural Network (CNN) for face spoofing detection. In \cite{wang2013face}, the pulse information and other spatial and temporal features learned by CNNs are fused together to boost the performance. In \cite{manjani2017detecting}, a multi-level deep dictionary learning based method was proposed especially for the silicone mask attacks. To improve the performance of CNNs, transfer learning based schemes have been adopted based on CNNs pretrained on ImageNet \cite{pinto2018counteracting}, VGG-Face \cite{parkhi2015deep} and GoogLeNet \cite{patel2016cross}. Motivated by the denoising algorithms, in \cite{jourabloo2018face} the spoofing patterns are treated as spoofing noise in the live face and extracted by a CNN architecture for classification.

However, due to the limited size of existing labeled data, the CNN models may be prone to over-fitting. To address this issue, methods can be classified into three categories. The first one is using the auxiliary information. In \cite{atoum2017face}, Atoum \textit{et al.} proposed a two-steam CNN based model, where one stream is responsible for patch-based anti-spoofing while another one is developed for depth-estimation. In addition to depth, remote Photoplethysmography (rPPG) signals have also been exploited by a CNN-RNN scheme from the raw videos in \cite{liu2018learning}. Considering the domain shift between different databases, domain adaptation based methods have been proposed to shrink the domain gap between samples in training and testing sets. In \cite{li2018unsupervised}, Li \textit{et al.} proposed an embedding function to map the image data into another space, such that the Maximum Mean Discrepancy (MMD) based loss can be optimized to evaluate the similarity of source and target domains. In recent work \cite{wang2019improving}, Wang \textit{et al.} utilized a generative adversarial network for domain adaptation based face spoofing detection. The shared embedding space by both the source and target domains can be learned when the discriminator cannot reliably predict whether a sample is from the source or target domain. However, due to the requirement of attack samples in the test database, domain adaptation based method may not be practical as it is difficult to acquire the spoofing images for any unseen device. The last category is domain generalization based methods. In \cite{li2018learning}, Li~\textit{et al.} utilized a 3D CNN model for the spatial-temporal information extraction. To reduce the domain shift among different domains, the regularization term is incorporated by minimizing the MMD. To learn a more generalized representation for face anti-spoofing, Tu~\textit{et al.} adopted the Total Pairwise Confusion (TPC) \cite{tu2019learning} loss for CNN training and moreover an identity based method was studied.
 
\begin{figure*}[t]
\begin{minipage}[b]{1\linewidth}
  \centering
  \centerline{\includegraphics[width=1\linewidth]{ 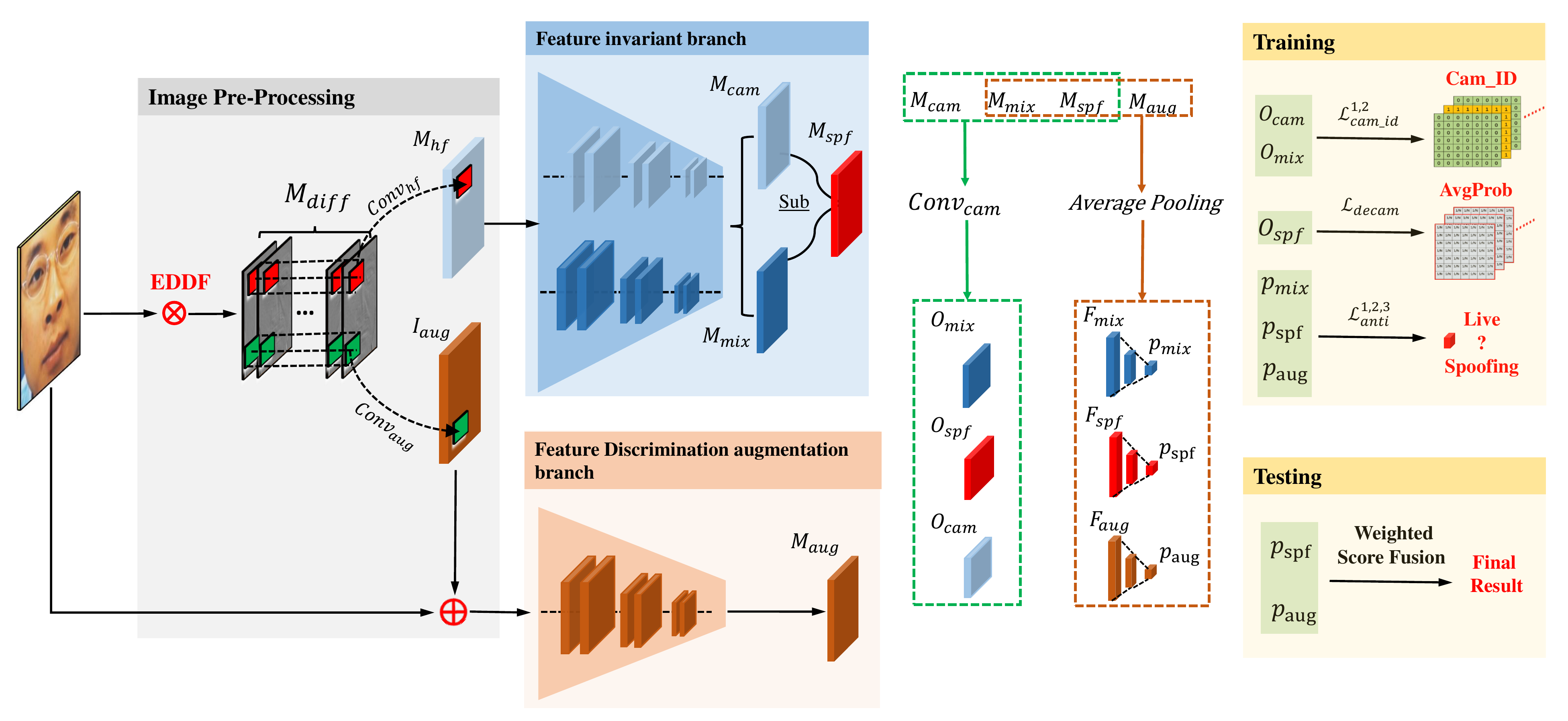}}
\end{minipage}
\caption{Illustration of the proposed framework which consists of two branches. In the feature invariant branch, the camera information is removed from the features that are specifically responsible for spoofing detection in HF domain. In the feature discrimination augmentation branch, we aim to extract more discriminative features based on a recomposition of the learned low/high-frequency components. Finally, the classification results of two branches are fused for the final prediction.}
\label{fig:3}
\end{figure*}

\section{The Proposed Scheme}

\subsection{Framework}

Generally speaking, the hardware and software processing in visual information acquisition leave unique traces in final images or videos~\cite{LiSource,Sutcu2007Improvements,chen2007digital}. These distinct fingerprints, which usually lie in the HF domain (e.g., sensor pattern noise),  imply unique camera information and exhibit strong invariance to the captured scene. Unfortunately, important clues in performing face spoofing such as the moiré pattern in replay attack or texture of artificial materials (paper, mask) also belong to the HF domain. As such, to obtain camera invariant spoofing features, the camera contamination causing the domain divergence between different cameras should be eliminated in a scientifically sound way. 

As illustrated in Fig.~\ref{fig:3}, we propose a two branch based model, in an effort to extract camera-free and spoofing specific features to achieve generalized face anti-spoofing. 
In the first branch, we focus on the HF information which contains abundant clues regarding camera and spoofing relevant features. Instead of performing metric learning to learn the camera irrelevant features, we treat the camera information as a factor varying the distribution of the spoofing features such that a feature decomposition scheme is proposed to align features captured from different cameras to pursue camera-invariant features learning. In particular, {{unlike   the conventional siamese CNN architecture \cite{zagoruyko2015learning} which is designed with shared weights  for  these  sub-networks,  we  propose  an architecture with the pseudo-siamese network, in which the two sub-networks share the same structure  while each sub-network will learn its own weights. More  specifically,  the  first sub-network  aims  for  camera  information  extraction,  and  the second  sub-network  targets  for  extracting  spoofing  features accompanied  with  camera  information.}} Due to the fact that two sub-networks share the same camera classifier, the first sub-network is expected to extract the same camera information existing in the second one. Then we utilize a feature decomposition scheme, with which the spoofing features can be independently extracted based on the obtained camera feature. However, straightforwardly using the first branch may limit the performance as only HF features are considered, while other spoofing clues including lighting, reflection \textit{etc.} tend to be neglected. In view of this, in the second branch, we extract discrimination augmented features based on the enhanced image recomposed from low/high-frequency signal to enhance the detection accuracy.
Finally, the detection results of the two branches are fused for final classification.

\subsection{Pre-processing of Input Images}

\begin{figure}[t]
\begin{minipage}[b]{0.9\linewidth}
  \centering
  \centerline{\includegraphics[width=1\linewidth]{ 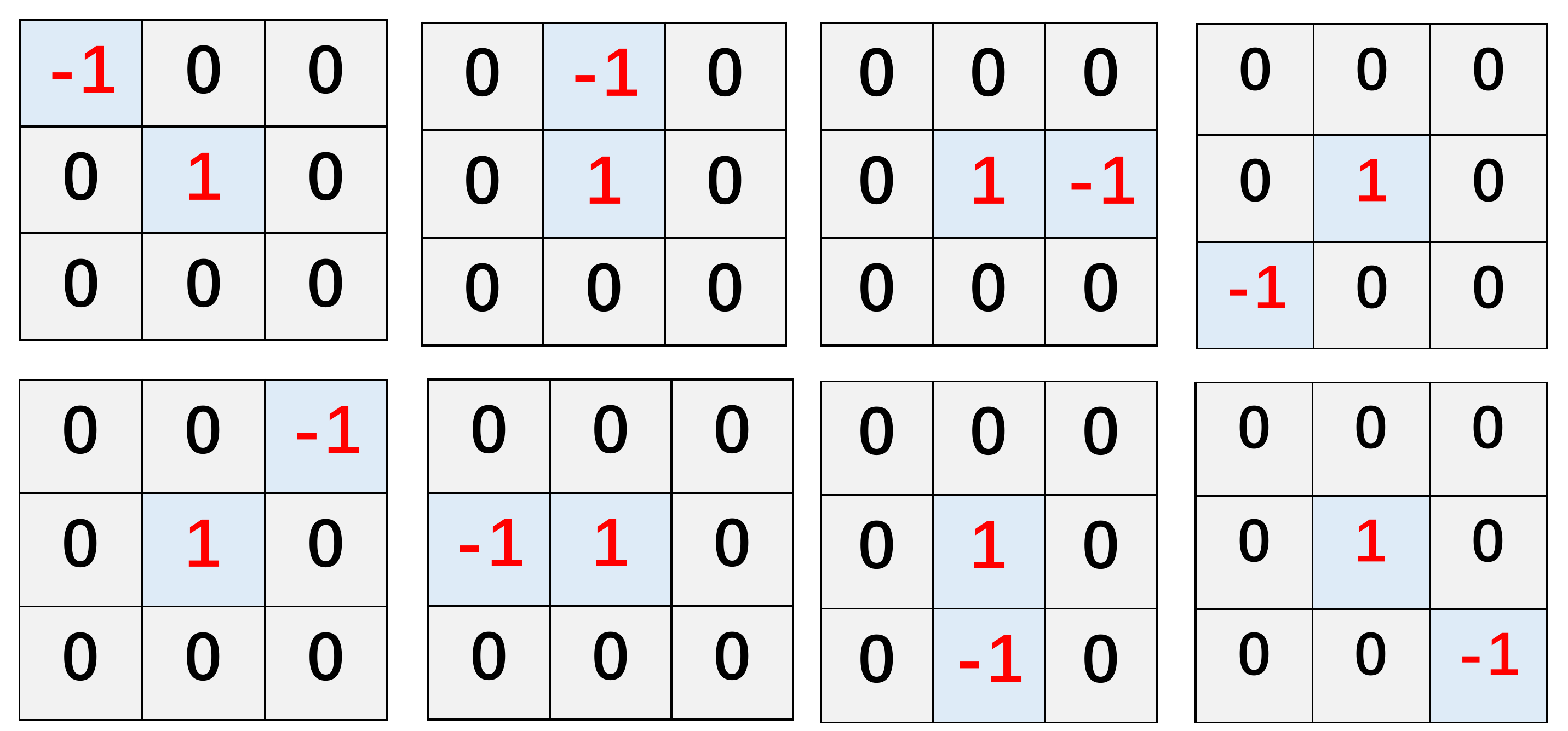}}
\end{minipage}
\caption{Illustration of the eight kernels in EDDF.}
\label{fig:4}
\end{figure}

Apparently, face anti-spoofing relying on textures aims to extract the unique features that could distinguish acquired genuine facial skin with the screen, paper, photo and mask. In analogous to the camera information, these clues are irrelevant to the face identity but closely relevant to the texture quality, motivating us to first focus on the HF domain. 
As shown in Fig.~\ref{fig:4}, the Eight-Direction Differential Filter-set (EDDF) is adopted for high-frequency information extraction. In particular, to eliminate the influence of the background, we first crop the face region from the original image using a face detection algorithm proposed in \cite{zhang2016joint}. Subsequently, we perform the filtering based on EDDF such that the fixed 2D convolution layers with eight $3\times3$ kernels on three channels of the input image can be obtained and the output 24 feature maps are denoted as $M_{diff}$. Then the $M_{diff}$ is further enhanced by enlarging the respective field adaptively with a multi-channel CNN layer of kernel size $5\times 5$, and the corresponding CNN layer is denoted as ${Conv_{hf}}$ in Fig.~\ref{fig:3}. {{ The EDDF is adopted based on the inspiration of steganalysis rich model (SRM) \cite{fridrich2012rich}, where the high-frequency components can be extracted by a union of many diverse submodels, leading to comprehensive representations of the high frequency information with eight corresponding high pass filters. The SRM has also been widely used for image manipulation detection \cite{zhou2018learning}  as well as camera filter array (CFA) patterns extraction \cite{goljan2015cfa}. Instead of using bunches of high pass filters, we employ the EDDF to provide the basic residual operations, then followed by ${Conv_{hf}}$ the useful high-frequency components can be extracted upon the output from EDDF adaptively.}} We denote the output feature maps as $M_{hf}$ and it will be treated as the input of a pseudo-siamese network for extracting spoofing specific features with camera information eliminated.  Instead of using only learned features extracted from the HF information,  the second branch learns to recompose the low/high frequency signal of the original image to augment spoofing clues for more discriminative feature learning. To this end, in this branch, the augmentation components $I_{aug}$ are learned by a three channel convolution layer ${Conv_{aug}}$ performed on $M_{diff}$. As such, the learned maps will be combined with the original image to generate an augmented image, serving as the input of the second branch. 




\subsection{Camera Invariant Feature Learning}

Given the HF domain input, the camera invariant feature learning aims to obtain the discriminative and generalized features based on feature level decomposition. 
More specifically, a pseudo-siamese network is adopted in an effort to remove camera variance from spoofing specific features. As illustrated in Fig.~\ref{fig:3}, in the first branch, the two sub-networks of the pseudo-siamese network share the same structure with three residual layers in ResNet-18 \cite{he2016deep}. However, they are individually trained towards different targets. 
In particular, the upper sub-network in the first branch is specifically designed to obtain camera relevant features. Given the fact that the camera information is determined by the camera type, the classification loss on camera types is adopted to guide the generation of the camera information. Herein, the learned feature maps at the third residual layer are denoted as $M_{cam}$. With $M_{cam}$, a camera classifier $Conv_{cam}$ is implemented based on a $\Phi$ channel CNN layer, where $\Phi$ is the number of categories of cameras in the training database. 
Considering the hard example mining, based on the original focal loss \cite{lin2017focal} we design a per-pixel multi-class focal loss  on output maps of last CNN layer $O_{cam}$ for camera type identification,
\begin{equation}\label{}
\mathcal{L}_{cam_{-} i d}^{1}=-\sum_{i=1}^{M} \sum_{j=1}^{N} \sum_{k=1}^{\Phi} y_{i,j, k}^{cam}\left(1-P_{i,j, k}^{cam}\right)^{\gamma} \log \left(P_{i,j, k}^{cam}\right),
\end{equation}
\begin{equation}\label{}
 P_{i,j,k}^{cam}=\frac{e^{h_{i,j,k}^{cam}}}{\sum_{l=1}^{\Phi} e^{h_{i,j,l}^{cam}}}.
\end{equation}
Herein, $M$ and $N$ indicate the spatial sizes of $O_{cam}$, and $y_{i,j,k}^{cam}$ is the binary label of camera type. In particular, when an element at location $(i,j)$ of $O_{cam}$ belongs to the camera type $k$, $y_{i,j,k}^{cam}=1$, and otherwise $y_{i,j,k}^{cam}=0$. $\gamma$ is a constant parameter for penalty adjustment on the easy examples. $P_{i,j,k}^{cam}$ denotes the prediction result from a softmax function in Eqn. (2), where $h_{i,j,k}^{cam}$ denotes the output of $k^{th}$ channel at spatial location $(i,j)$ in $O_{cam}$. The reason to adopt the pixel-wise loss to form the constraint is that the camera information should be shared and maintained identical among local patches sampled from an image. Moreover, each element of the feature $O_{cam}$  corresponds to a patch of the input images. When we choose to enforce the constraint on these elements, it amounts to sample training patches from the image repeatedly, which largely improves the diversity of data and alleviates the over-fitting problem.


The bottom sub-network of the first branch aims to learn the spoofing specific features. In analogous with the upper sub-network, the learned feature maps at the third residual layer are denoted as $M_{mix}$, as shown in Fig.~\ref{fig:3}. Following $M_{mix}$, the $F_{mix}$ is acquired by an average pooling layer and two fully connected layers produce the prediction on spoofing with a softmax function. The binary (live \textit{vs.} spoofing) focal loss is adopted to persuade the model to well discriminate hard examples and increase the distance between genuine and spoofing samples, such that more discriminative spoofing features can be learned. As such, it is formulated as follows,
\begin{equation}\label{}
\begin{aligned}
\mathcal{L}^{1}_{\text {anti}}=&-\alpha_{1}(y^{mix}(1-p^{mix})^\gamma \log (p^{mix})) \\
 &+\alpha_{2}((1-y^{mix})({p^{mix}})^\gamma \log (1-p^{mix})),
\end{aligned}
\end{equation}
where $y^{mix}$ is the ground truth, ($y^{mix}$ = 0 for spoofing and $y^{mix}$ = 1 for live) and $p^{mix}$ is predicted probability of the live sample. $\alpha_{1}$, $\alpha_{2}$ and $\gamma $ are constant parameters to control the balance between hard and easy samples. The spoofing features extracted from $M_{mix}$ are mixed with camera information. As such, to further eliminate the camera variance, the camera classification loss is also imposed on the generated output Maps $O_{mix}$ from $M_{mix}$, to drive the training of the bottom sub-network based on the classification loss. 
\begin{equation}\label{}
\mathcal{L}_{cam_{-} i d}^{2}=-\sum_{i=1}^{M} \sum_{j=1}^{N} \sum_{k=1}^{\Phi} y_{i,j, k}^{mix}\left(1-P_{i,j, k}^{mix}\right)^{\gamma} \log \left(P_{i,j, k}^{mix}\right).
\end{equation}


Herein, our aim is to construct the approximation of the latent camera invariant features conditioned on the explicitly learned $M_{mix}$ and  $M_{cam}$ based on decomposition, with a strong prior that the resultant features have a strong capability in discriminating spoofing patterns while no information regarding camera types is implied. 
To this end, we formulate a rational feature level decomposition model by establishing the relationship among $M_{mix}$, $M_{cam}$ and the desired camera invariant HF spoofing features $M_{spf}$.  As the feature $M_{spf}$ expected to acquire is independent with the camera type, we assume that the $M_{mix}$ is a combination of $M_{spf}$ and the camera information denoted as $M_{cam}^{mix}$, leading to a feature decomposition approach,
\begin{equation}\label{}
\textit{$M_{mix} = M_{cam}^{mix} + M_{spf} $.}
\end{equation}
As $M_{cam}^{mix} $ shares the same camera classifiers with $M_{cam} $, we have,
\begin{equation}\label{}
\textit{$M_{cam}^{mix} = M_{cam} $.}
\end{equation}
As such, the camera invariant HF spoofing map can be acquired by,
\begin{equation}\label{}
\textit{$M_{spf} = M_{mix}-M_{cam} $.}
\end{equation}

In general, the desired feature $M_{spf}$ should not be able to identify the camera type while equipped with strong capability in detecting spoofing.  
In particular, the loss that confuses the camera type \textit{decam} is defined as follows,
\begin{equation}\label{}
\mathcal{L}_{\text {decam}}=-\sum_{i=1}^{M} \sum_{j=1}^{N} \sum_{k=1}^{\Phi} y_{i,j,k}^{spf} \log \left(P_{i,j,k}^{spf}\right),
\end{equation}
\begin{equation}
 \mathrm{y}_{\mathrm{i}, \mathrm{j},\mathrm{k}}^{spf}=\frac{1}{\Phi}.
\end{equation}
As such, the predicted probability of each camera type is desired to be equal to the average, indicating that the $M_{spf}$ loses the capability in discriminating the camera for camera-invariant feature learning. 

Given $M_{spf}$, the pooled feature map $F_{spf}$ can be acquired with an average pooling layer and the face liveness prediction is achieved with two fully connected layers. For training the network,  the binary (live \textit{vs.} spoofing) focal loss is adopted again, 
\begin{equation}\label{}
\begin{aligned}
\mathcal{L}^{2}_{\text {anti}}=&-\alpha_{1}(y^{spf}(1-p^{spf})^\gamma \log (p^{spf})) \\
 &+\alpha_{2}((1-y^{spf})({p^{spf}})^\gamma \log (1-p^{spf})),
\end{aligned}
\end{equation}
where $y^{spf}$ is the ground truth and $p^{spf}$ is predicted probability of the face liveness. 
The prediction result based on $F_{spf}$ from the first branch is used to detect the spoofing.

{It is worth mentioning that in this branch we adopt the camera type as the side information for camera invariant feature learning. In the training phase, we divide the face images in terms of camera types in the training set, then the pixel-wise multi-class cross-entropy loss is imposed for learning and extracting the camera feature. The per-pixel based loss is adopted with the assumption that the camera information existing in an image should be invariant to the acquired scene. Moreover, the number of camera categories for classification varies with the used training set. However, we do not attempt to employ the classification results in the testing phase, as totally different camera types from the training set may form the testing set. Instead, we only use the features learned in the second to last layer as the extracted camera information. Although the cameras in the testing set are unseen during training, the camera-specific information can also be successfully extracted. Subsequently, the camera feature will be decomposed from the learned mixed spoofing feature in the second sub-network, by which the camera invariant feature can be finally learned in the first branch. }

\subsection{Feature Discrimination Capability Augmentation}
The first branch aims to obtain camera invariant spoofing features in the HF domain, which will inevitably ignore certain useful information in other frequency ranges, \textit{i.e.}, color, camera reflectance. To comprehensively obtain features extracted from both high and low frequency domains for spoofing detection, the second branch is specifically learned.
In the second branch, to emphasize discriminative spoofing clues, we adopt the augmented image $I_{aug}$ generated in the image pre-processing phase as the input. Again, after the last residual layer we acquire the augmented feature map $M_{aug}$ and an average pooling layer is adopted to reduce the spatial dimension. We denote the output features as $F_{aug}$ and subsequently two fully connected layers with a softmax function predict the final result. 
The binary focal loss is also adopted here,
\begin{equation}\label{}
\begin{aligned}
\mathcal{L}^{3}_{\text {anti}}=&-\alpha_{1}(y^{aug}(1-p^{aug})^\gamma \log (p^{aug})) \\
 &+\alpha_{2}((1-y^{aug})({p^{aug}})^\gamma \log (1-p^{aug})),
\end{aligned}
\end{equation}
where $y^{aug}$ is the ground truth and $p^{aug}$ is the predicted probability of the face liveness. 

\subsection{Spoofing Detection}
In summary, six loss functions have been defined in the two branches and the total loss for training our network is a weighted sum of these loss functions,
\begin{equation}
\begin{aligned}
Loss=& \lambda_{1}\left(\mathcal{L}_{c a m_{-} i d}^{1}+\mathcal{L}_{c a m_{-} i d}^{2}\right) \\
&+\lambda_{2}\left(\mathcal{L}_{a n i}^{1}+\mathcal{L}_{a n i}^{2}+\mathcal{L}_{a n i}^{3}\right)+\lambda_{3} \mathcal{L}_{d e c a m},
\end{aligned}
\end{equation}
where $ \lambda_{1} $, $ \lambda_{2} $, $ \lambda_{3} $ are the weighting factors.
In the testing phase, we use a weighted score to fuse the results from the two branches,
\begin{equation}
P=\frac{P(F_{spf})+\lambda_{4} P(F_{aug})}{1+\lambda_{4}},
\end{equation}
where $\lambda_{4}$ is the empirically set fusion score weight. $P(F_{spf})$ and  $P(F_{aug})$ are the predicted probabilities of face liveness from the two branches. 

\section{Experimental Results}
\label{sec:typestyle}

\begin{figure}[ht]
\begin{center}
\includegraphics[width=1.0\textwidth]{ 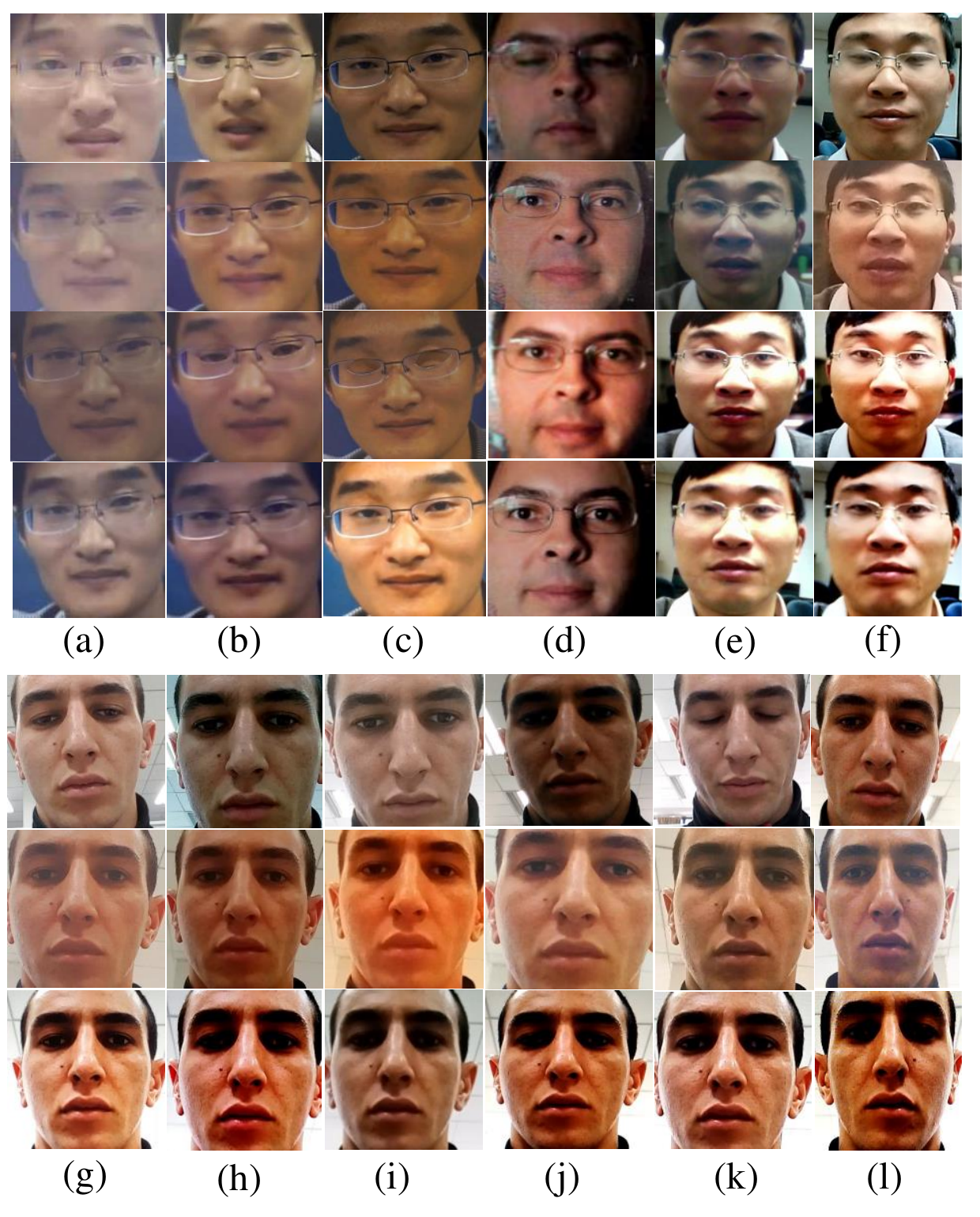}
\end{center}
\caption{{Illustrations of face samples in the four datasets. (a), (b) and (c) are images sampled from three camera types (low quality, normal quality and high quality) in the CASIA-FASD database \cite{zhang2012face}. From top to down are the live samples and spoofing samples generated by printed photo, photo with eye cut and video replay. (d) are images sampled from Replay-Attack database  \cite{chingovska2012effectiveness}. From top to down are the live sample and spoofing samples generated by print photo, screen photo and video replay. (e) and (f) are images sampled from two types of cameras (MacBook Air and Google Nexus 5) in MSU-MFSD database \cite{wen2015face}. From top to down are the live samples and spoofing samples generated by printed photo and video replay with iPad and iPhones respectively. (g)-(l) are images sampled from six mobile cameras {(Samsung Galaxy S6 edge, HTC Desire EYE, MEIZU X5, ASUS Zenfone Selfie, Sony XPERIA C5 Ultra Dual and OPPO N3)} in Oulu-NPU \cite{boulkenafet2017oulu} database. From top to down are the live samples and spoofing samples generated by print photo and video replay.}}
\label{fig:sample}
\end{figure}

\subsection{Datasets}

We evaluate our model on four face anti-spoofing datasets: CASIA-FASD \cite{zhang2012face}, Replay-Attack \cite{chingovska2012effectiveness}, Oulu-NPU \cite{boulkenafet2017oulu} and MSU-MFSD \cite{wen2015face}. The descriptions of the four databases are shown in Table 1.

\textbf{\textit{CASIA-FASD:}} The CASIA-FASD database was published in 2012. In this database, 50 subjects are included. For each subject, 3 live and 9 fake video clips are provided. There are three attack types in this database including warped photo attack, cut photo attack and video replay attack. The cameras can be categorized into three different quality levels including low quality (web camera), normal quality (web camera) and high quality (Sony NEX-5). Among the 600 video clips, 240 are used for the training set and the rest are used for testing.

\textbf{\textit{Replay-Attack:}} This database also contains 50 subjects. The total number of videos is 12000, consisting of 360 videos (60 real-accesses and 300 attacks) in the training set, 360 videos (60 real-accesses and 300 attacks) in the development set and 480 videos (80 real-accesses and 400 attacks) in the testing set. Two types of attack  including print attack and replay attack are performed in this database. For the print attack, the face images are printed on a high resolution A4 paper. For the replay attack, two devices are adopted, including iPhone 3GS (with resolution $ 480\times320 $) and first generation iPad (with resolution $ 1024\times768 $). Different illumination conditions are studied in this database. The first is a controlled condition where a fluorescent lamp is used for lighting with a uniform background. The other condition is adverse, and only day-light is used in the background. All videos in this database are recorded by a webcam of a MacBook.

 \textbf{\textit{Oulu-NPU:}} This database was published in 2017 and it consists of 4950 video clips. Those videos are divided into three sets: 1) training set, which contains 360 real and 1440 attack videos of 20 subjects. 2) development set, which contains 270 real and 1080 attack videos of 15 subjects. 3) testing set, which contains 360 real and 1400 attack videos of 20 subjects. Six cameras are used for recording in this database. The attack types implemented in this database are print and video-replay, generated by two types of printers and two types of display devices, respectively. Four testing protocols are considered to evaluate the environmental condition variations, spoofing mediums variation, camera style variations and fusion of all above challenges. 

\textbf{\textit{MSU-MFSD:}} Two different resolutions of cameras are considered in this database: 1) the built-in camera in the MacBook Air 13 with the resolution of $ 640\times 480$, 2) the front-facing camera in the Google Nexus 5 with resolution of $ 720\times 480$. The database contains 35 subjects and 280 videos in total. Two attack types are performed including printed photo and video replay. For the printed photo attack, the Canon 550D camera is used for high resolution ($ 5184\times 3456$) face image acquisition and the photo will be printed on an A3 paper with an HP color printer. Two cameras are used for replay attacks including SLR camera and iPhone 5S back-facing camera.

{The sample face images of the four datasets, including CASIA-FASD, Replay-Attack, MSU-MFSD and Oulu-NPU, are shown in Fig.~\ref{fig:sample}.  Both live samples and spoofing samples generated by different attack types are presented. It should be noted that there is no overlapping on the subjects between training set and testing set, implying that we test our model on unseen subjects in both intra-dataset and cross-dataset settings.}

\begin{table*}
\begin{floatrow}
\scalebox{0.8}{
  \centering
  \caption{Descriptions of the four databases.}
\begin{tabular}{m{9.51em}<{\centering}|c <{\centering}|m{4.125em} <{\centering}|m{6.625em} <{\centering}|c <{\centering}|m{10.25em} <{\centering}|m{6.25em} <{\centering}}
\hline
\textbf{Database} & \textbf{Year} & \textbf{Number of Identities} & \textbf{Genuine/Attack Samples} & \textbf{Attack Types} & \textbf{Capture Cameras} & \textbf{Display Cameras} \bigstrut\\
\hline
\hline
CASIA-FASD \cite{zhang2012face} & 2012  & 50    & 150/450 & 1 Print, 2 Replay & Low-quality webcam, normal-quality webcam, SonyNEX-5 & iPad \bigstrut\\
\hline
REPLAY-ATTACK \cite{chingovska2012effectiveness} & 2012  & 50    & 200/1000 & 1 Print, 2 Replay & Macbook webcam & iPhone 3GS, iPad  \bigstrut\\
\hline
MSU-MFSD \cite{wen2015face} & 2015  & 35    & 110/330 & 1 Print, 2 Replay & Macbook air webcam, nexus 5 & iPad Air, iPhone 5S \bigstrut\\
\hline
OULU-NPU \cite{boulkenafet2017oulu} & 2017  & 55    & 1980/3960 & 2 Print, 2 Replay & Samsung Galaxy S6, edge oppo\_n3, htc\_desire\_eye, meizu\_x5, asus xenfone, sony\_xperia\_c5  & Dell1905FP,\newline{}MacBook Retina \bigstrut\\
\hline
\end{tabular}%

  \label{tab:addlabel}%
}
\end{floatrow}
\end{table*}%

\subsection{Implementation Details} 
We implement our model by PyTorch. For each database, we averagely select 30 frames from each videos. To eliminate the background variations, we crop the face region from each frame by a face detection algorithm \cite{zhang2016joint}. As our proposed method is camera based method, we divide the database based on camera types. We divide the CASIA-FASD into 3 sub-databases and MSU-MFSD into 2 sub-databases based on the camera types. As there is only one single camera in Replay-Attack, this database is not divided. The cropped face images are resized to $ 224\times 224$. Considering the imbalance of positive and negative samples, we sample the live faces and spoofing faces at a ratio 1:1 in a batch during training. We also use data augmentation for better generalization capacity. The augmentation includes horizontal and vertical flipping with a probability of 0.5, random rotation within 15 degrees as well as color jitter. The color jitter is implemented based on the \textit{Transforms} module in Pytorch. In Table~\uppercase\expandafter{\romannumeral2}, we show the layer-wise network design of our proposed method. It is worth mentioning that we replace batch normalize layers with group normalization layers~\cite{wu2018group}, and the group number is set to 32 in each residual layer to make our network more stable.

The batch size in the training phase is 32 and we adopt Adam optimizer for optimization. The learning rate begins at 0.004 for the first 20k steps and reduces by 0.2 after every 10k steps. As the Oulu-NPU database contains much more samples compared with other databases, we set the learning rate to be 0.004 for the first 30k steps, which further reduces by 0.2 after every 20k steps. We set the parameters
$\alpha_{1}, \alpha_{2}, \gamma, \lambda_{1}, \lambda_{2},\lambda_{3}, \lambda_4 $ be 0.5, 1.0, 4.0, 0.005, 5.0, 0.1, 0.7 for all experiments. Five metrics are adopted for performance evaluations, including:

1) Equal Error Rate (EER). We use EER for intra-database evaluation and the development set is used for EER threshold selection.

2) Half Total Error Rate (HTER). HTER is the average of False Acceptance Rate (FAR) and False Rejection Rate (FRR). It is defined as follows:
\begin{equation}
H T E R=\frac{F A R+F R R}{2}.
\end{equation}

3) Attack Presentation Classification Error Rate (APCER). It is defined as follows:
\begin{equation}
A P C E R_{P A I}=\frac{1}{N_{P A I}} \sum_{i=1}^{N_{P A I}}\left(1-R_{e s_{i}}\right),
\end{equation}
where $ N_{P A I} $ is the number of attack attempts of certain Presentation Attack Instruments (PAI). The value $ R_{e s_{i}} $ will be `1’ if an attempt is predicted as `attack' and 0 as `live’. From the definition, we can find that the ACER only considers the worst performance of different scenarios, which means it penalizes approaches that only perform well on certain types of attacks. 

4) BonaFide Presentation Classification Error Rate (BPCER):
\begin{equation}
B P C E R=\frac{\sum_{i=1}^{N_{B F}}\left(R_{e s_{i}}\right)}{N_{BF}},
\end{equation}
where $ N_{B F} $ is the total number of the Bonafide presentation times. 

5) Average Classification Error Rate (ACER). It is the average of APCER and BPCER:
\begin{equation}
A C E R=\frac{A P C E R+B P C E R}{2}.
\end{equation}
The APCER, BPCER and ACER are the standardized ISO/IEC 30107-3 metrics in \cite{BiometricPresentation}.

\subsection{ Comparison with the State-of-the-Art Methods }
To demonstrate the high generalization capability of our method, we firstly perform cross-database experiments on above mentioned four databases and compare our results with the state-of-the-art methods. We adopt seven cross-test settings in this experiment, including: CASIA-FASD $ \rightarrow$ Replay-Attack, CASIA-FASD $ \rightarrow$ MSU-MFSD, MSU-MFSD $\rightarrow$ Replay-Attack, MSU-MFSD $\rightarrow$ CASIA-FASD, (CASIA-FASD \& MSU-MFSD) $\rightarrow$ Replay-Attack, MSU-MFSD \& Replay-Attack $\rightarrow$ CASIA-FASD, CASIA-FASD \& Replay-Attack $\rightarrow$ MSU-MFSD. To simplify, we denote these seven settings as C $\rightarrow$ R, C $\rightarrow$ M, M $\rightarrow$ R, M $\rightarrow$ C, (C \& M ) $\rightarrow$ R, (M \& R) $\rightarrow$ C, (C\& R) $\rightarrow$ M. The R $\rightarrow$ C and R $\rightarrow$ M settings are not involved, as the camera classification cannot be performed considering that only one camera type exists in the Replay-Attack database. The comparison results are reported in Table ~\uppercase\expandafter{\romannumeral3}.

{As can be seen, we compare our method with 14 classical methods, including both hand-crafted feature based methods and CNN based methods.  It should be mentioned that for CNN based methods, classical  CNN  models  are  involved,  including  original  ResNet-18 \cite{he2016deep} for  binary  classification, CNN \cite{yang2014learn}, ResNet18+TripletLoss provided in \cite{wang2019improving}, DeepPixel \cite{george2019deep} method and Auxiliary \cite{liu2018learning} with depth map only which is denoted as Auxiliary (depth only).} From Table ~\uppercase\expandafter{\romannumeral3}, we can find our method achieve the lowest HTER in most cross-database settings except for M $\rightarrow$ C and C $\rightarrow$ M. Regarding the M $\rightarrow$ C and C $\rightarrow$ M settings, our method achieves the second best HTER results which are competitive with state-of-the-art models (27.3\% \textit{vs.} 24.3\% and 17.5\% \textit{vs.} 14.0\%). Moreover, for C $\rightarrow$ R, we achieve more than 10\% lower in terms of HTER than the  state-of-the-art methods ``Auxiliary'', which demonstrate the effectiveness of our method. It is worth noting that although the \textit{TripletLoss} is utilized in \cite{wang2019improving} aiming to reduce the intra-class distance and enlarge the inter-class distance, it still tends to be over-fitting and cannot generalize well on different databases. It demonstrates that our feature decomposition scheme is more suitable in the face anti-spoofing task as some meaningful spoofing clues may be discarded due to the distance restricted by \textit{TripletLoss}.
{In Table ~\uppercase\expandafter{\romannumeral3}, we also show the HTER results of the two branches with different settings. 
It is obvious that although only using the second branch cannot achieve the desired performance, it provides the compulsory  information for the first branch and the results can be further improved by fusion of the two branches, which reveals that the results of the  two  branches  can be  effectively leveraged  by our fusion strategy.}

\begin{figure}[t]
\begin{minipage}[b]{0.9\linewidth}
  \centering
  \centerline{\includegraphics[width=1\linewidth]{ 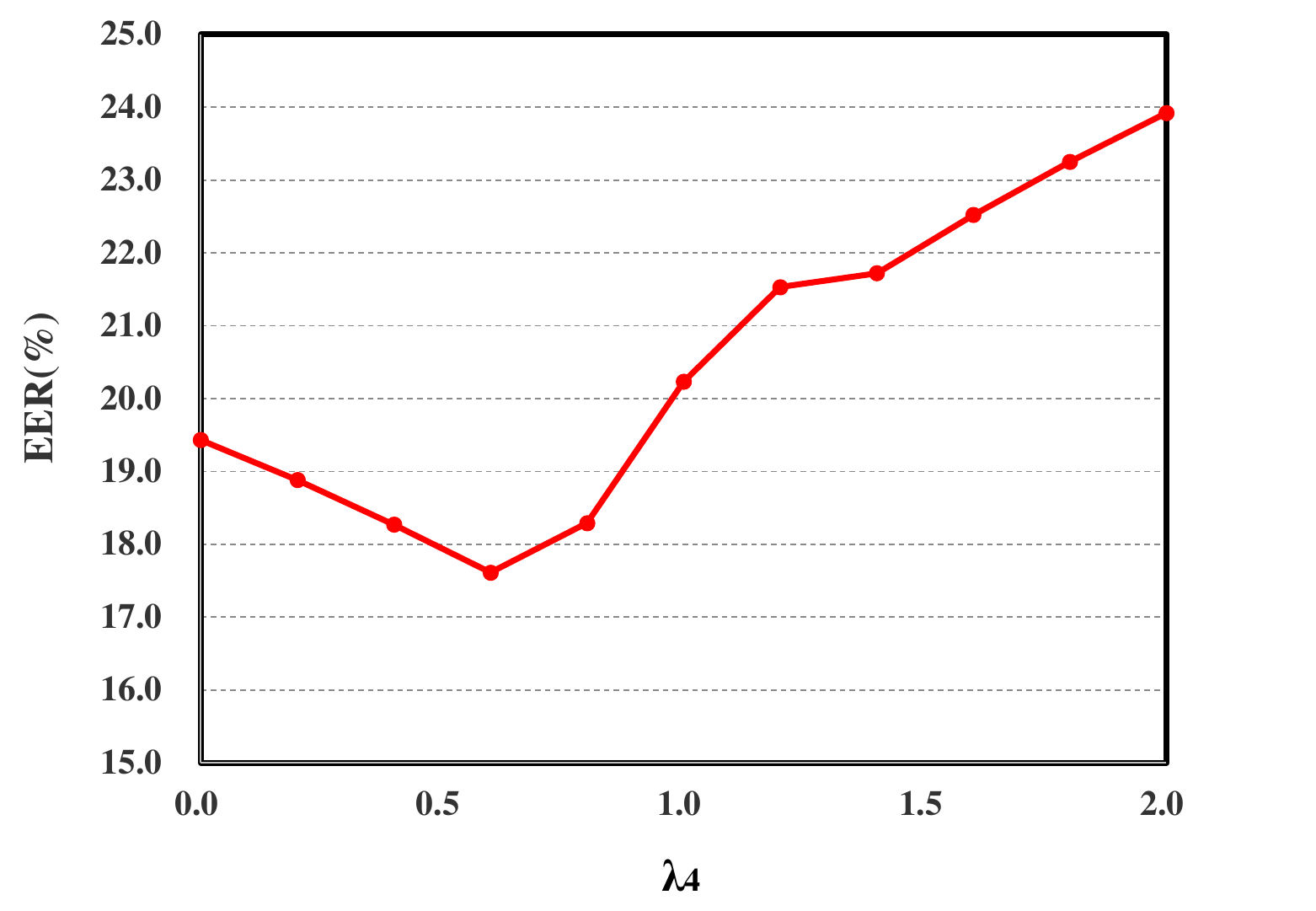}}
\end{minipage}
\caption{ The EER(\%) results influenced by the fusion weighting parameter $ \lambda_{4} $ in Eqn. (13) on cross-database (C $ \rightarrow$ R) setting.}
\label{fig:8}
\end{figure}

\begin{figure}[t]
\begin{minipage}[b]{0.85\linewidth}
  \centering
  \centerline{\includegraphics[width=1\linewidth]{ 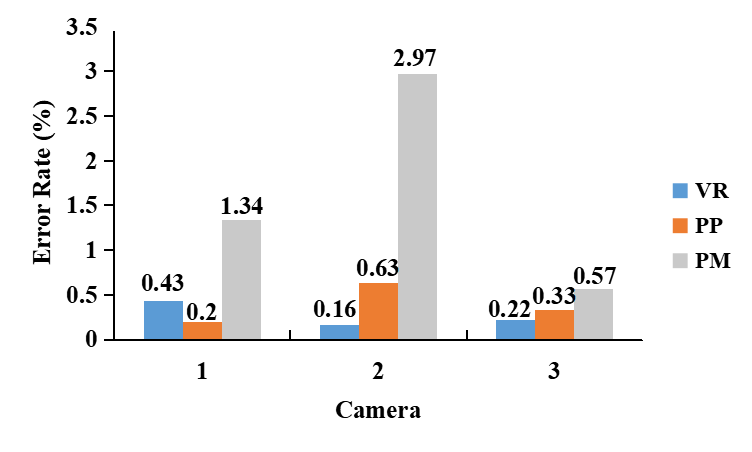}}
\end{minipage}
\caption{ The error rate of the false accepted attack types on the {CASIA-FASD} database. `VR' means Video Replay, `PP' indicates the printed photo and `PM' means a mask made by the printed photo with the regions of eyes and mouth been cut.}
\label{fig:6}
\end{figure}

\begin{table*}
\begin{floatrow}
\scalebox{0.9}{
  \centering
  \caption{Architecture of the network in the proposed method.}
\begin{tabular}{cc|lrr}
\hline
\multicolumn{2}{c|}{Layer name} & \multicolumn{1}{c|}{Input size} & \multicolumn{1}{c|}{[Kernel size, Channels]} & \multicolumn{1}{c}{Output size} \bigstrut\\
\hline
\hline
\multicolumn{5}{c}{\textbf{Feature Invariant Branch}} \bigstrut\\
\hline
\multicolumn{2}{c|}{EDDF} & \multicolumn{1}{c}{224×224} & \multicolumn{1}{c}{[3×3, 24]} & \multicolumn{1}{c}{224×224} \bigstrut[t]\\
\multicolumn{2}{c|}{$Conv_{hf}$} & \multicolumn{1}{c}{224×224} & \multicolumn{1}{c}{[5×5, 64]} & \multicolumn{1}{c}{224×224} \bigstrut[b]\\
\hline
\multicolumn{1}{c|}{\multirow{10}[2]{*}{Residual Blocks }} & \multicolumn{1}{c|}{Conv} & \multicolumn{1}{c}{224×224} & \multicolumn{1}{c}{[7×7, 64], stride=2} & \multicolumn{1}{c}{112×112} \bigstrut[t]\\
\multicolumn{1}{c|}{} & \multicolumn{1}{c|}{Pooling} & \multicolumn{1}{c}{112×112} & \multicolumn{1}{c}{3×3 Maxpooling, stride=2} & \multicolumn{1}{c}{56×56} \\
\multicolumn{1}{c|}{} & \multicolumn{1}{c|}{\multirow{3}[0]{*}{Residual Bloack1}} & \multicolumn{1}{c}{\multirow{3}[0]{*}{56×56}} & \multicolumn{1}{c}{\multirow{3}[0]{*}{$\left[\begin{array}{l}
3 \times 3,128 \\
3 \times 3,128
\end{array}\right] \times 2$}} & \multicolumn{1}{c}{\multirow{3}[0]{*}{56×56}} \\
\multicolumn{1}{c|}{} &       &       &       &  \\
\multicolumn{1}{c|}{} &       &       &       &  \\
\multicolumn{1}{c|}{} & \multicolumn{1}{c|}{\multirow{3}[0]{*}{Residual Bloack2}} & \multicolumn{1}{c}{\multirow{3}[0]{*}{56×56}} & \multicolumn{1}{c}{\multirow{3}[0]{*}{$\left[\begin{array}{l}
3 \times 3,256 \\
3 \times 3,256
\end{array}\right] \times 2$}} & \multicolumn{1}{c}{\multirow{3}[0]{*}{28×28}} \\
\multicolumn{1}{c|}{} &       &       &       &  \\
\multicolumn{1}{c|}{} &       &       &       &  \\
\multicolumn{1}{c|}{} & \multicolumn{1}{c|}{\multirow{2}[1]{*}{Residual Bloack3}} & \multicolumn{1}{c}{\multirow{2}[1]{*}{28×28}} & \multicolumn{1}{c}{\multirow{2}[1]{*}{$\left[\begin{array}{l}
3 \times 3,512 \\
3 \times 3,512
\end{array}\right] \times 2$}} & \multicolumn{1}{c}{\multirow{2}[1]{*}{14×14}} \\
\multicolumn{1}{c|}{} &       &       &       &  \bigstrut[b]\\
\hline
\multicolumn{2}{c|}{$Conv_{cam}$} & \multicolumn{1}{c}{14×14} & \multicolumn{1}{c}{[3×3, Number of Cameras]} & \multicolumn{1}{c}{14×14} \bigstrut\\
\hline
\hline
\multicolumn{5}{c}{\textbf{Feature Discrimination Augmentation Branch}} \bigstrut\\
\hline
\multicolumn{2}{c|}{$Conv_{aug}$} & \multicolumn{1}{c}{224×224} & \multicolumn{1}{c}{[3×3,3]} & \multicolumn{1}{c}{224×224} \bigstrut\\
\hline
\multicolumn{1}{c|}{\multirow{10}[2]{*}{Residual Blocks }} & \multicolumn{1}{c|}{Conv} & \multicolumn{1}{c}{224×224} & \multicolumn{1}{c}{[7×7,64], stride=2} & \multicolumn{1}{c}{112×112} \bigstrut[t]\\
\multicolumn{1}{c|}{} & \multicolumn{1}{c|}{Pooling} & \multicolumn{1}{c}{112×112} & \multicolumn{1}{c}{3×3 Maxpooling, stride=2} & \multicolumn{1}{c}{56×56} \\
\multicolumn{1}{c|}{} & \multicolumn{1}{c|}{\multirow{3}[0]{*}{Residual Bloack1}} & \multicolumn{1}{c}{\multirow{3}[0]{*}{56×56}} & \multicolumn{1}{c}{\multirow{3}[0]{*}{$\left[\begin{array}{l}
3 \times 3,128 \\
3 \times 3,128
\end{array}\right] \times 2$}} & \multicolumn{1}{c}{\multirow{3}[0]{*}{56×56}} \\
\multicolumn{1}{c|}{} &       &       &       &  \\
\multicolumn{1}{c|}{} &       &       &       &  \\
\multicolumn{1}{c|}{} & \multicolumn{1}{c|}{\multirow{3}[0]{*}{Residual Bloack2}} & \multicolumn{1}{c}{\multirow{3}[0]{*}{56×56}} & \multicolumn{1}{c}{\multirow{3}[0]{*}{$\left[\begin{array}{l}
3 \times 3,256 \\
3 \times 3,256
\end{array}\right] \times 2$}} & \multicolumn{1}{c}{\multirow{3}[0]{*}{28×28}} \\
\multicolumn{1}{c|}{} &       &       &       &  \\
\multicolumn{1}{c|}{} &       &       &       &  \\
\multicolumn{1}{c|}{} & \multicolumn{1}{c|}{\multirow{2}[1]{*}{Residual Bloack3}} & \multicolumn{1}{c}{\multirow{2}[1]{*}{28×28}} & \multicolumn{1}{c}{\multirow{2}[1]{*}{$\left[\begin{array}{l}
3 \times 3,512 \\
3 \times 3,512
\end{array}\right] \times 2$}} & \multicolumn{1}{c}{\multirow{2}[1]{*}{14×14}} \\
\multicolumn{1}{c|}{} &       &       &       &  \bigstrut[b]\\

\hline
\multicolumn{2}{c|}{Average Pooling} & \multicolumn{1}{c}{14×14} &  & \multicolumn{1}{c}{1×1} \bigstrut\\

\hline
\multicolumn{2}{c|}{\multirow{4}[4]{*}{Binary Classification}} & \multicolumn{1}{c}{ $F_{mix}$ } & \multicolumn{1}{c}{ $F_{spf}$ } &  \multicolumn{1}{c}{$F_{aug}$}  \bigstrut\\

\cline{3-5}\multicolumn{2}{c|}{} &  \multicolumn{1}{c}{FC1 (512, 128)} & \multicolumn{1}{c}{FC1 (512, 128)} & \multicolumn{1}{c}{FC1 (512, 128)} \bigstrut[t]\\
\multicolumn{2}{c|}{} & \multicolumn{1}{c}{Relu}  & \multicolumn{1}{c}{Relu}  & \multicolumn{1}{c}{Relu} \\
\multicolumn{2}{c|}{} &\multicolumn{1}{c}{ FC2 (128, 2)} & \multicolumn{1}{c}{FC2 (128, 2)} & \multicolumn{1}{c}{FC2 (128, 2)} \bigstrut[b]\\
\hline
\end{tabular}%
  \label{tab:addlabel}%
}
\end{floatrow}
\end{table*}%
Furthermore, we also evaluate the performance influenced by different fusion weighting parameter $ \lambda_{4} $ on cross-database testing (C $\rightarrow$ R). As shown in Fig.~\ref{fig:8}, we can find that the performance will drop if the value of $ \lambda_{4} $ is set to an inappropriate value (extremely lager or small). The best value of $ \lambda_{4} $ is around 0.5 which is half of the weight (1.0) of the first branch. It demonstrates that the HF feature invariant spoofing clues learned from the first branch are more important than the spoofing features learned by the second branch when the testing data exhibit large distribution divergence against the data in the training database. 
\begin{table*}
\begin{floatrow}
\scalebox{1}{
  \centering
  \caption{Cross-test results (HTER \%) on CASIA-FASD, Replay-Attack, and MSU-MFSD. ``-'' indicates that the corresponding
result is unavailable. The numbers in bold are the best results.}
\begin{tabular}{c|ccccccc}
\hline
\textbf{Method} & \textbf{C $\rightarrow$ R} & \textbf{M $\rightarrow$ R} & \textbf{M $\rightarrow$ C} & \textbf{C $\rightarrow$ M} & \textbf{C\&M $\rightarrow$ R} & \textbf{C\&R $\rightarrow$ M} & \textbf{M\&R $\rightarrow$ C} \bigstrut[b]\\
\hline
\hline
LBP \cite{chingovska2012effectiveness} & 47.0    & 45.5  &-       &-       &-       &-       &-  \bigstrut[t]\\

LBP-TOP \cite{de2014face} & 49.7  & 46.5  &-       &-       &-       &-       &-  \\
Motion \cite{anjos2011counter} & 50.2  &-       &-       &-       &--       &-       &-   \\
CNN \cite{yang2014learn} & 48.5  & 37.1  & 37.8  & 26.3  & 29.3  & 21.2  & 37.2 \\
Color LBP \cite{boulkenafet2018generalization}  & 37.9  & 44.8  & 45.7  & 21.0    &-       &-      &-  \\
Color Tex. \cite{boulkenafet2018generalization} & 30.3  & 33.9  & 46.0    & 20.4  &-       &-       &-   \\
Color SURF \cite{boulkenafet2018generalization} & 26.9  & 29.7  & \textbf{24.3}  & 19.1  &-       &-       &-    \\
ResNet18 \cite{he2016deep}  & 47.0  & 47.9  & 45.8  & 36.4  & 35.3  & 27.6  & 40.7 \\
ResNet18+TripletLoss \cite{wang2019improving} & 43.3  & 11.5  & 37.8  & \textbf{14.0}  &27.3       &23.8       &43.3 \\
Auxiliary \cite{liu2018learning} & 27.6  &-       &-       &-       &-       &-       &-   \\
Auxiliary (depth only)\cite{liu2018learning}  & 29.1  & 26.7  & 44.5  & 36.1  & 29.7  & 16.6  & 35.5   \\
DeepPixel \cite{george2019deep}  & 41.5  & 21.9  & 36.0  & 40.3  & 38.9  & 21.6  & 33.1   \\
De-Spoof \cite{jourabloo2018face} & 28.5  &-       &-       &-       &-       &-       &-   \bigstrut[b]\\
\hline
Invariant Branch (Ours)  & 19.4  & 25.8  & 32.6  & 21.6  & 24.6  & 15.0  & 31.2  \\
Augmentation Branch (Ours)  & 31.2  & 28.1  & 36.3  & 31.9  & 29.3  & 18.6  & 37.8  \\
Fusion  (Ours) & \textbf{17.6}  & \textbf{21.7}  & 27.3  & 17.5  & \textbf{21.3}  & \textbf{14.8}  & \textbf{32.3} \bigstrut\\
\hline
\end{tabular}%
 \label{tab:addlabel}%
}
\end{floatrow}
\end{table*}%
To further verify the effectiveness of our method, we also perform intra-database experiments on {CASIA-FASD} database. The results are shown in Table~\uppercase\expandafter{\romannumeral4}. From Table~\uppercase\expandafter{\romannumeral4}, we can find our method could achieve the best EER result (0.89\%), demonstrating our method does not sacrifice the accuracy of intra-database even though our main target is to improve the generalization capability. 
For further analysis of the performance of our method under different attack types, we show the error rates of three attack types: Video Replay (VR), Print Photo (PP) and cut Photo Mask (PM) in Fig.~\ref{fig:6}. We can find that our method achieves lower EER when the attack type is `VR' and `PP'. However, although the `PM' also uses printed paper as an attack medium, our method cannot detect this attack type perfectly. The potential reason is the live face parts (eyes and mouth) existing in the input image may cause the contamination in the spoofing features. 

\begin{figure}[t]
\begin{minipage}[b]{0.8\linewidth}
  \centering
  \centerline{\includegraphics[width=1\linewidth]{ 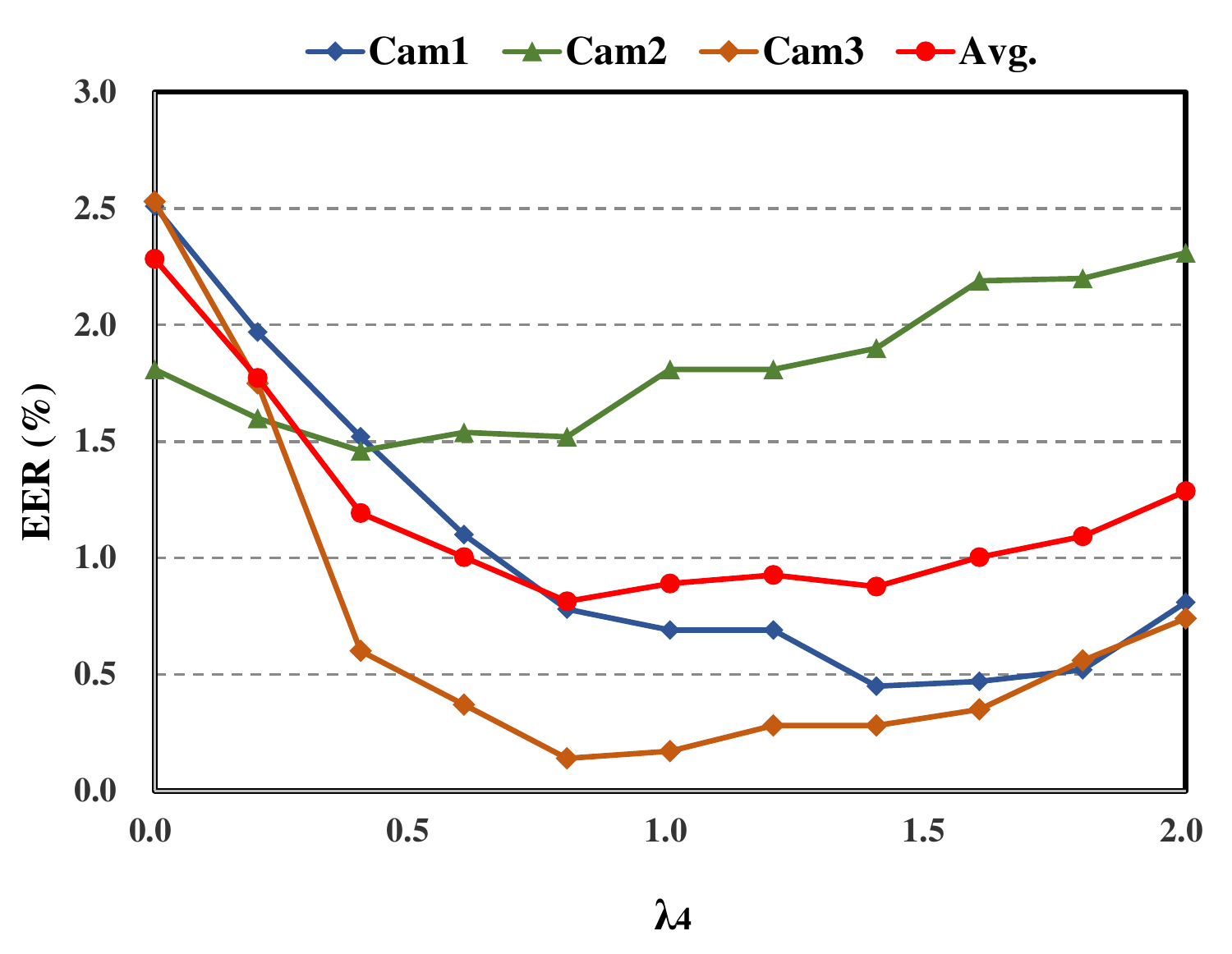}}
\end{minipage}
\caption{Variations of the EER(\%) results influenced by the fusion weight parameter $ \lambda_{4} $ in Eqn. (13) on  intra-database (CASIA-FASD) setting.}
\label{fig:7}
\end{figure}

Analogously, we investigate the influence of the parameter $ \lambda_{4} $ of Eqn. (13) on the performance, we set different weightings regarding the fusion of the two branches, and the results are shown in Fig.~\ref{fig:7}. From Fig.~\ref{fig:7}, we can find a similar phenomenon that the EER results will be higher if $ \lambda_{4} $ is set too small or too lager, and the best performance can be achieved when $ \lambda_{4} $ is set to 0.8. This indicates that the two branches in our model are all necessary, and provides useful evidence on the effectiveness of our fusion strategy. 

Then we conduct an intra-database experiment on Oulu-NPU. Four protocols are provided in the OULU-NPU database. Protocol 1 is used to evaluate the generalization capability in different environments. Protocol 2 is designed for performance evaluation on unseen attack types including video replay and printed photo. Protocol 3 is the most relevant to our method, which is used to evaluate how a model can be generalized to unseen camera types. Six camera types are contained in this protocol, in which, five selected camera styles are used for training and the rest one for testing. Protocol 4 is the most challenging protocol, as different environments, attack mediums and cameras are all considered in this protocol. we report our experimental results in Table ~\uppercase\expandafter{\romannumeral5}. 

Form the table, we observe that our method can achieve the best ACER in Protocol 3 with the lowest variance in terms of different cameras which indicates that our method can be generalized well to the camera modules. Moreover, our method achieves superior results compared with the “GRADIANT” \cite{boulkenafet2017competition} which only uses binary labels for supervision. In Protocols 2 and 4, we achieve competitive performance compared with “Auxiliary” or the “DeSpoofing” method which also uses the auxiliary depth information for training. It is worth noting that our method achieves the lowest variance in Protocol 3 as well as Protocol 4, demonstrating the promising robustness of our method. 

\begin{table}
\begin{floatrow}
\scalebox{0.85}{
  \centering
  \caption{Intra-test results on {CASIA-FASD}. The number in bold are the best results.}
\begin{tabular}{c<{\centering}|c<{\centering}}
\hline
Method & EER(\%) \bigstrut\\
\hline
\hline
IQA \cite{galbally2014face} & 32.4 \bigstrut\\

Motion \cite{anjos2011counter} & 26.6 \bigstrut\\

LBP \cite{chingovska2012effectiveness} & 18.2 \bigstrut\\

LBP-TOP \cite{de2014face} & 10.0 \bigstrut\\

LBP+SVM \cite{boulkenafet2016face2} & 4.9 \bigstrut\\
CNN \cite{yang2014learn} & 7.4 \bigstrut\\

MSR-RESNET \cite{chen2019attention}& 3.1 \bigstrut\\

Color SURF  \cite{boulkenafet2018generalization}& 2.2 \bigstrut\\

ResNet18 \cite{he2016deep} & 5.2\bigstrut\\
DeepPixel \cite{george2019deep}& 2.6\bigstrut\\
ResNet18+TripletLoss \cite{wang2019improving} & 1.4\bigstrut\\
Auxiliary (depth only)\cite{liu2018learning}& 1.3\bigstrut\\

\hline
CamInva Branch (Ours)  & 2.3    \\
Augmentation Branch (Ours)  & 2.9  \\
Fusion  (Ours) & \textbf{0.89} \bigstrut\\
\hline
\end{tabular}%
  \label{tab:addlabel}%
}

\end{floatrow}
\end{table}%

\begin{table}
\begin{floatrow}
\scalebox{0.7}{
  \centering
  \caption{Experimental results of the four protocols on the OULU-NPU database.}
\begin{tabular}{c<{\centering}|m{7.96em}<{\centering}|m{5.04em}<{\centering}|m{4.71em}<{\centering}|m{4em}<{\centering}}
\hline
\multicolumn{1}{c|}{\multirow{2}[2]{*}{Prot.}} & \multirow{2}[2]{*}{Methods} & \multirow{2}[2]{*}{APCER(\%)} & \multirow{2}[2]{*}{BPCER(\%)} & \multirow{2}[2]{*}{ACER(\%)} \bigstrut[t]\\
      & \multicolumn{1}{c|}{} & \multicolumn{1}{c|}{} & \multicolumn{1}{c|}{} & \multicolumn{1}{c}{} \bigstrut[b]\\
\hline
\hline
\multirow{5}[10]{*}{1} & CPqD \cite{boulkenafet2017competition} & \multicolumn{1}{c|}{2.9} & \multicolumn{1}{c|}{10.8} & \multicolumn{1}{c}{6.9} \bigstrut\\
     & GRADIANT \cite{boulkenafet2017competition} & \multicolumn{1}{c|}{1.3} & \multicolumn{1}{c|}{12.5} & \multicolumn{1}{c}{6.9} \bigstrut\\
      & Auxiliary \cite{liu2018learning} & \multicolumn{1}{c|}{1.6} & \multicolumn{1}{c|}{\textbf{1.6}} & \multicolumn{1}{c}{1.6} \bigstrut\\
      & DeSpoofing \cite{jourabloo2018face} & \multicolumn{1}{c|}{\textbf{1.2}} & \multicolumn{1}{c|}{1.7} & \multicolumn{1}{c}{\textbf{1.5}} \bigstrut\\
     & Ours  & \multicolumn{1}{c|}{3.8} & \multicolumn{1}{c|}{2.9} & \multicolumn{1}{c}{3.4} \bigstrut\\
\hline
\multirow{5}[10]{*}{2} & MixedFASNet \cite{boulkenafet2017competition} & \multicolumn{1}{c|}{9.7} & \multicolumn{1}{c|}{2.5} & \multicolumn{1}{c}{6.1} \bigstrut\\
     & GRADIANT \cite{boulkenafet2017competition} & \multicolumn{1}{c|}{3.1} & \multicolumn{1}{c|}{1.9} & \multicolumn{1}{c}{2.5} \bigstrut\\
      & Auxiliary \cite{liu2018learning} & \multicolumn{1}{c|}{\textbf{2.7}} & \multicolumn{1}{c|}{2.7} & \multicolumn{1}{c}{2.7} \bigstrut\\
     & DeSpoofing \cite{jourabloo2018face} & \multicolumn{1}{c|}{4.2} & \multicolumn{1}{c|}{4.4} & \multicolumn{1}{c}{4.3} \bigstrut\\
     & Ours  & \multicolumn{1}{c|}{3.6} & \multicolumn{1}{c|}{\textbf{1.2}} & \multicolumn{1}{c}{\textbf{2.4}} \bigstrut\\
\hline
\multirow{5}[10]{*}{3} & MixedFASNet \cite{boulkenafet2017competition} & 5.3±6.7 & 7.8±5.5 & 6.5±4.6 \bigstrut\\
      & GRADIANT \cite{boulkenafet2017competition}& \textbf{2.6}±3.9 & 5.0±5.3  & 3.8±2.4 \bigstrut\\
      & Auxiliary \cite{liu2018learning} & 2.7±1.3 & 3.1±1.7 & 2.9±1.5 \bigstrut\\
    & DeSpoofing \cite{jourabloo2018face} & 4.0±1.8  & 3.8±1.2 & 3.6±1.6 \bigstrut\\
     & Ours  & 3.8±1.3 & \textbf{1.1}±1.1 & \textbf{2.5}±0.8 \bigstrut\\
\hline
\multirow{5}[10]{*}{4} & Massy HNU \cite{boulkenafet2017competition} & 35.8±35.3 & 8.3±4.1 & 22.1±17.6 \bigstrut\\
     & GRADIANT \cite{boulkenafet2017competition} & \textbf{5.0}±4.5 & 15.0±7.1 & 10.0±5.0 \bigstrut\\
     & Auxiliary \cite{liu2018learning}& 9.3±5.6 & 10.4±6.0 & 9.5±6.0 \bigstrut\\
     & DeSpoofing \cite{jourabloo2018face}& 5.1±6.3 & \textbf{6.1}±5.1 & \textbf{5.6}±5.7 \bigstrut\\
     & Ours  & 5.9±3.3 & 6.3±4.7 & 6.1±4.1 \bigstrut\\
\hline
\end{tabular}%
  \label{tab:addlabel}%
}
\end{floatrow}
\end{table}%

\subsection{Ablation Study} 

In this subsection, to reveal the functionalities of different modules in the proposed method, we perform the ablation analysis. The experiments are conducted both on intra-database ({CASIA-FASD} database) and cross-database (training on CASIA-FASD and testing on Replay-Attack) settings. The results are shown in Table~\uppercase\expandafter{\romannumeral6} in which the feature invariant branch is denoted as the first branch and the feature discrimination augmentation branch is denoted as the second branch. Moreover, $ Cam1 $, $ Cam2 $ and $ Cam3 $ correspond to low quality, normal quality and high quality cameras in the CASIA-FASD dataset. To identify the effectiveness of the two branches, we show the experiment results of the two branches respectively without fusion and a T-SNE visualization of the learned feature in the first branch is shown in Fig.~2(b). Subsequently, we remove EDDF at each branch to evaluate the functionality of EDDF. More specifically, when evaluating each branch without EDDF, we adopt the same residual architecture in each branch except for the input image which has not been processed by the EDDF, and the number of channels of the following CNN layer is three. Another important factor is the feature decomposition operation. To verify its importance, we ablate the camera classification sub-network in the first branch such that the results of the second sub-network and the second branch are fused for comparison. We denote the results as \textit{Fusion w/o CamID} in Table~\uppercase\expandafter{\romannumeral6}.
\begin{table}
\begin{floatrow}
\scalebox{0.9}{
  \centering
  \caption{Ablation studies of our method on the CASIA database and Repaly-Attack.}
\begin{tabular}{m{4.75 em} <{\centering}|m{5.25em}|m{3.75em}<{\centering}|m{5.75em}<{\centering}}
\hline
\multirow{2}[4]{*}{Methos} & \multicolumn{2}{c|}{Intra Testing ( EER\% )} & Cross Testing ( HTER\% ) \bigstrut\\
\cline{2-4}      & \multicolumn{2}{c|}{{CASIA-FASD} } & {CASIA-FASD}  $\rightarrow $ Replay \bigstrut\\
\hline
\hline
1st branch & Cam1:  2.5\newline{}Cam2:  1.8\newline{}Cam3:  2.6 & Avg.: 2.3 & 19.4 \bigstrut\\
\hline
1st branch w/o EDDF & Cam1: 2.9\newline{}Cam2: 2.4\newline{}Cam3: 3.1 & Avg.: 2.8 & 24.8 \bigstrut\\
\hline
2nd branch & Cam1: 2.1\newline{}Cam2: 4.9\newline{}Cam3: 1.6 & Avg.: 2.9 & 31.2 \bigstrut\\
\hline
2nd branch w/o EDDF & Cam1: 3.3\newline{}Cam2: 7.0\newline{}Cam3: 2.5 & Avg.: 4.3 & 32.1 \bigstrut\\
\hline
Fusion  w/o CamID & Cam1: 2.4\newline{}Cam2: 1.2\newline{}Cam3: 3.1 & Avg.: 2.2 & 29.3 \bigstrut\\
\hline
Fusion & Cam1: 0.7\newline{}Cam2: 1.8\newline{}Cam3: 0.2 & Avg.: 0.9 & 17.6 \bigstrut\\
\hline
\end{tabular}%

  \label{tab:addlabel}%
}
\end{floatrow}
\end{table}%

From the EER results on {CASIA-FASD}  intra-database settings in Table~\uppercase\expandafter{\romannumeral6}, we can find that the variance of detection results of each camera is significantly reduced in the first branch compared with the second branch. It is demonstrated that more generalized spoofing features can be learned in the first branch. In addition, the performance of the first branch on $ Cam1 $ and $ Cam3 $ is degraded compared with the second branch, due to the fact that only the HF feature is utilized in the first branch. However, promising performance improvement (31.2 \textit{vs.} 19.4) by the first branch when testing on cross-database (training on {CASIA-FASD} and testing on Replay) has been observed. This phenomenon is reasonable, as more database-dependent information (\textit{e.g.}, environment, lighting) is used in the second branch, and the information will cause the learned features to tend to be over-fitting on the training database such that a negative effect when testing on a different database has been observed. Moreover, we find that the ablation of EDDF in the first branch will cause the performance drop on both intra-database and cross-database settings. It is demonstrated that the EDDF can make this branch pay attention to the HF component when the input image is pre-processed by EDDF. Similar results can be observed when the EDDF is ablated in the second branch. With EDDF, lower EER and HTER values can be acquired, as more discriminative parts can be augmented before being processed by CNN layers. As such, it can be concluded that the EDDF is necessary for both branches. From Table~\uppercase\expandafter{\romannumeral6}, we can find that the PAD performance will be largely improved when we fuse the results of two branches, and outperform each individual branch. This implies that the learned features of each branch are compulsory and the results of the two branches are balanced by the weighted fusion strategy.

\subsection{Feature Visualizations} 

\begin{figure*}[t]
\begin{minipage}[b]{0.8\linewidth}
  \centering
  \centerline{\includegraphics[width=1\linewidth]{ 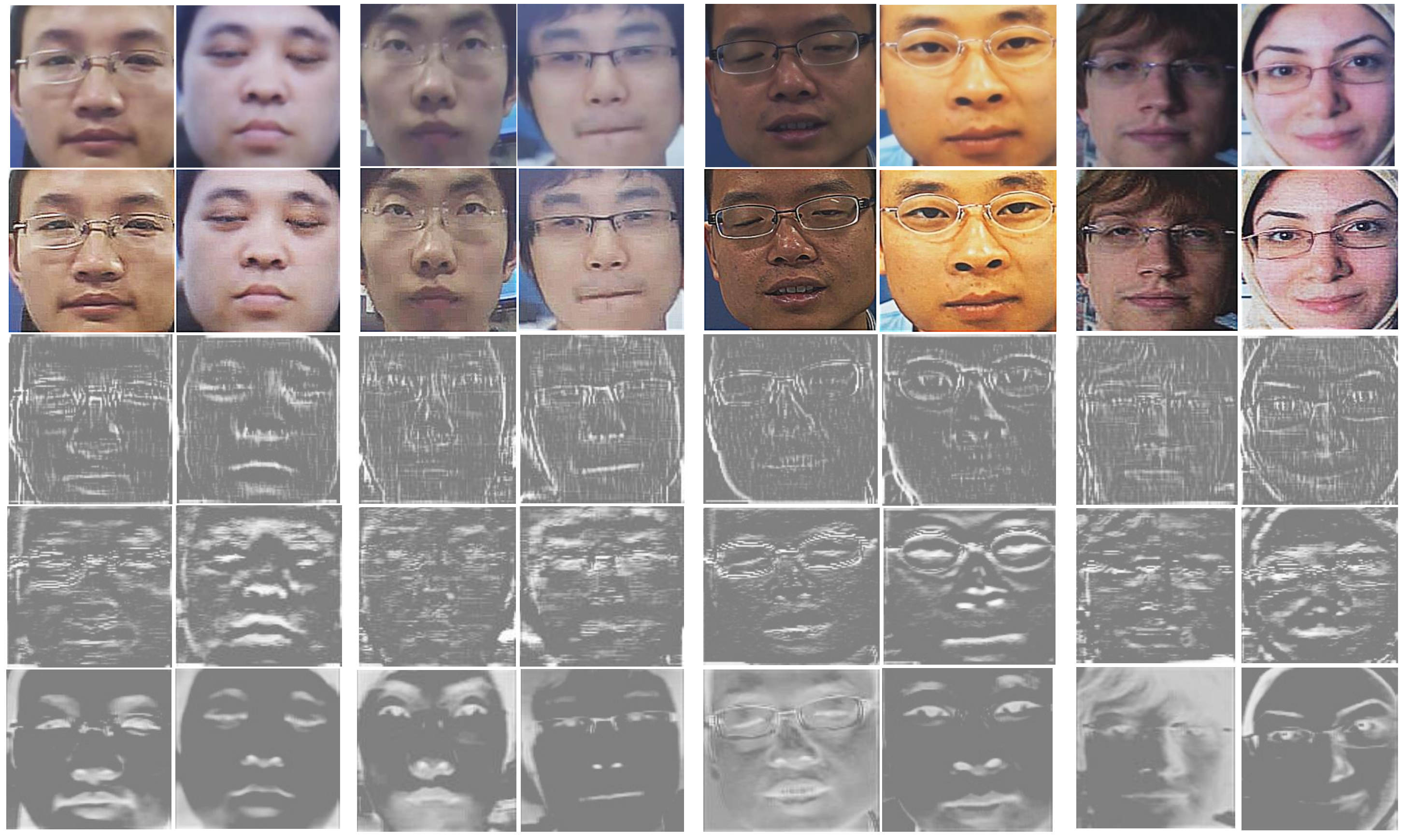}}
\end{minipage}
\caption{Visualization results of the learned feature maps. The model is trained on {CASIA-FASD} database. In the first row, from left to right are the input live and spoofing face images sampled from CASIA-FASD database (including three camera types: low quality camera, normal quality camera and high quality camera) and Replay Attack database. The second row shows the images augmented in the second branch. The sampled feature maps of the two subnetworks in the ``Feature Invariant Branch" are shown in the following two rows respectively and the last row presents one of the features learned in the ``Feature Discrimination Augmentation Branch".}
\label{fig:9}
\end{figure*}

To better understand the learned spoofing features of the two branches in our proposed methods, we train the model using {CASIA-FASD} database and visualize the feature maps of each sub-network. Moreover, to identify which components will be augmented in the second branch, we also visualize the input map in the second branch. The results are shown in Fig.~\ref{fig:9}.

As we can see, in the second row, the contrast and details of input images are augmented. Taking the augmented image as the input in the second branch, the last row shows the learned feature maps, from which we can find a large difference exists between the feature maps of live and spoofing face images, especially in the regions of the nose and mouth. The third row shows the maps learned for camera type classification. In this row, although the face images in the same camera type are different (live \textit{vs.} spoofing), similar camera patterns can still be extracted and the patterns are varied by camera types. In the fourth row, the spoofing feature maps learned in the second sub-network are shown. From the images, we can find more noise exists in the spoofing face image compared with live samples. However, the noise patterns are different among different cameras as the camera information mixed with the spoofing patterns causes the domain gap among different cameras.

  {\subsection{Investigations on the Robustness of Unknown Camera Feature Extraction}
In the feature invariant branch of  our model, we use the camera feature extracted in the first sub-network as the side information to guide the model to remove the camera information incorporated in the spoofing detection feature, leading to the camera invariant feature extraction for spoofing detection. In this sense, the camera features extracted in the testing set should be more specific and different from the camera features extracted in the training set (as the camera type in the testing set is not identical with that of the training set). To demonstrate such assumption, we provide the visualization of cameras features extracted from training set and testing set in Fig.~\ref{fig:tsne} (equal number of samples of each dataset are randomly selected for  better visualization).  More specifically, the number of camera types in training set ranges from 2 to 5.

\begin{figure*}[ht]
\begin{center}
\includegraphics[width=0.75\textwidth]{ 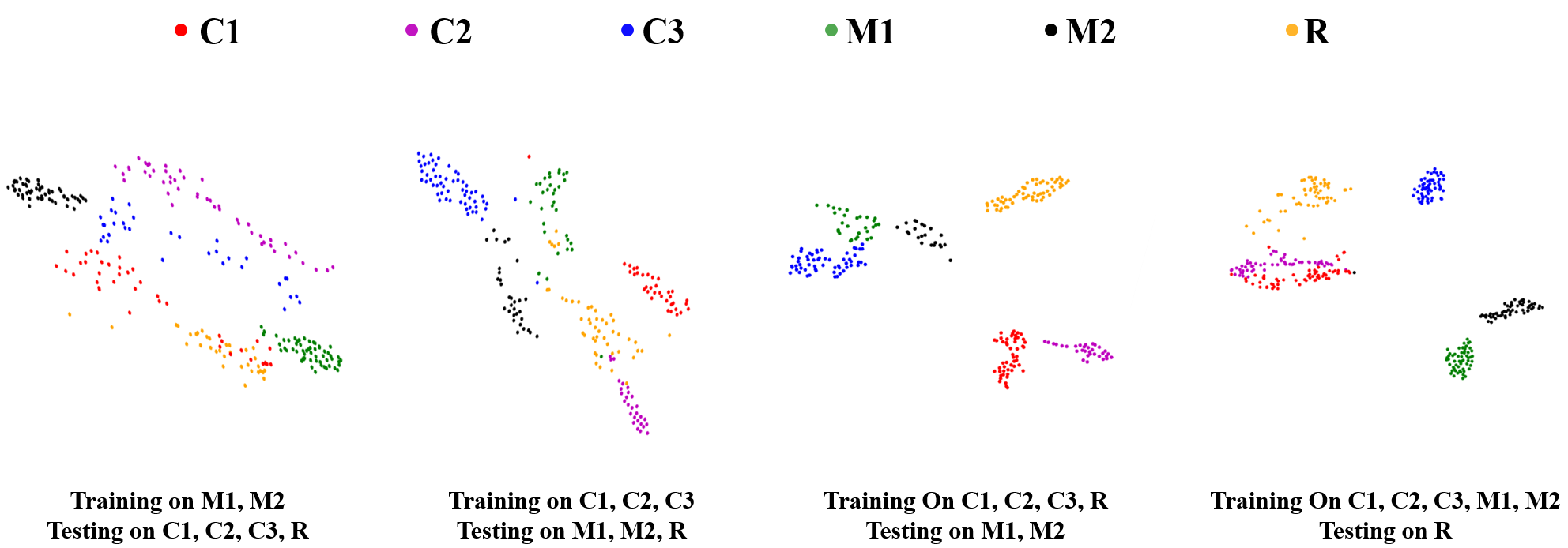}
\caption{ T-SNE [18] visualization of the camera features. C1, C2, C3, M1, M2 and R represent the six different camera types in CASIA-FASD, MSU-MFSD and Replay-Attack datasets.}
\label{fig:tsne}
\end{center}
\end{figure*}

As shown in Fig.~\ref{fig:tsne}, we use different numbers of camera types for camera classifier training and the results show that our camera classifier can extract the camera-specific features of unseen cameras in the testing set even the number of camera type in the training set is limited (\textit{e.g.} 2 or 3). Besides, we can also find that the camera classification capability can be improved (corresponding to more compact of each camera feature) when the number of camera type for training increases. However, even the discriminative camera features can be efficiently extracted by the camera classifier,   {it cannot be fully guaranteed that the features do not carry other forms of noise which are not considered in the training data. To improve the robustness of camera feature extraction, we further explore the robustness of unknown camera feature extraction and conduct more studies on the probability based unknown camera classification with a self-attention based strategy for the feature refining.}

\textit{\textbf{Unknown camera classification.}}
  {In this step, we study the ``\((N + 1)\)”  camera classifier, where \(N\) is the number of camera categories in the training set. In the testing phase, if a sample is classified into  \((N+1)-th\) category, the sample is further treated as an image captured by an unknown camera. Towards the implementation of the \(N+1\) camera classifier, we adopt the softmax probability distribution as category descriptor which is widely used in the filed of Out-of-Distribution Detection (OOD) detection~\cite{hendrycks2016baseline,yu2019unsupervised,mandal2019out}. } To be specific, we assume the camera classifier is trained by cameras in \(N\) categories and tested on a testing set with \(N+M\) categories, where \(M\) is the number of unknown camera types. We first show the average softmax probability of each camera type for  testing in Fig.~\ref{fig:pplot}. Furthermore, we also visualize the probability distribution on intra-dataset for comparisons in Fig.~\ref{fig:pinta}. From the figures, we can observe that the probability distribution of cameras types in the training set has significant differences with the distribution of unknown camera types. Inspired by this observation, to transfer the trained classifier into an `` \((N + 1)\)” classifier, the distribution descriptor is created. More specifically, for each sample $x^{t}$ in the testing set, we use the difference between the maximum classification probability $p^{\max1st}$ and second maximum probability $p^{\max2nd}$ that $x^{t}$ is predicted among \(N\) categories to infer the probability that the $x^{t}$  belongs to the \((N+1)-th\) category,
\begin{equation}\label{} 
p^t\left(y^{t} = N+1\right)=1-\frac{p^{\max 1st}\left(x^{t}\right)-p^{\max 2nd}\left(x^{t}\right)}{\tau},
\end{equation}
{where $y^{t}$ is the predicted camera category of $x^{t}$ and the parameter $\tau$ can be calculated from the training set. To be specific, for each sample  $x^{s }$ in the training set, if its  maximum classification probability $p^{\max 1st}\left(x^{s}\right)>0.6$, this sample is selected and the corresponding distribution descriptor $\tau\left(x^{s}\right)$ is obtained as follows:
\begin{equation}\label{} 
\tau\left(x^{s}\right) =\frac{p^{\max 1st}\left(x^{s}\right)-p^{\max 2nd}\left(x^{s}\right)}{1-p^{\max1st }\left(x^{s}\right)}.
\end{equation}
The final $\tau $ is selected as the minimum value among the descriptors of all selected samples.}
Then the normalized probability that $x^{t}$ belongs to category m (m $\in \{1,2,3...N+1\}$) can be acquired as follows:
\begin{equation}\label{} 
p^{nor}\left(y^{t} = m\right)=\frac{p^t\left(y^{t} =m\right)}{\sum_{j=1}^{N+1} p^t\left(y^{t} =j\right)}.
\end{equation}
To demonstrate the performance of our ``\(N + 1\)” classifier in the testing set, we average the probability that each sample is classified into \((N+1)-th\) category by camera type. The results are shown in Fig.~\ref{fig:np1}. Besides, the intra-testing results on CASIA-FASD dataset are also visualized in Fig.~\ref{fig:n1inta} for comparisons. It is apparent that the unknown camera is classified into the \((N+1)-th\) category with a high probability ($\ge$ 0.85) and the intra-class cameras are rejected by the \((N+1)-th\) category. This reveals the effectiveness of the ``\(N + 1\)” classifier.

\begin{figure*}[ht]
\begin{center}
\includegraphics[width=1.0\textwidth]{ 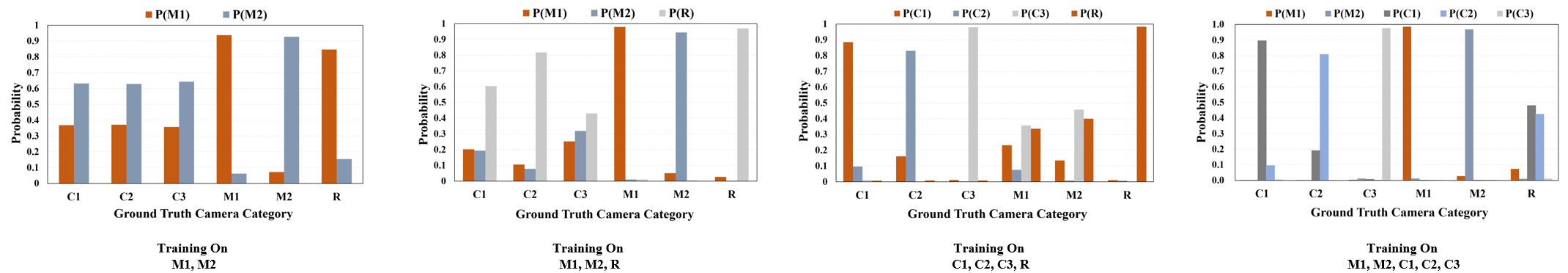}
\caption{Visualization of the average probability of 
camera-specific samples in the \(N\) category classification, where \(N\) is the number of camera categories in the training set. The datasets for training is annotated under each sub-figure and all samples in six datasets are used for testing. The probability \(p(\cdot)\) corresponds to the probability that the testing sample is classified into the corresponding category. 
}
\label{fig:pplot}
\end{center}
\end{figure*}

\begin{figure}[ht]
\begin{center}
\includegraphics[width=0.65\textwidth]{ 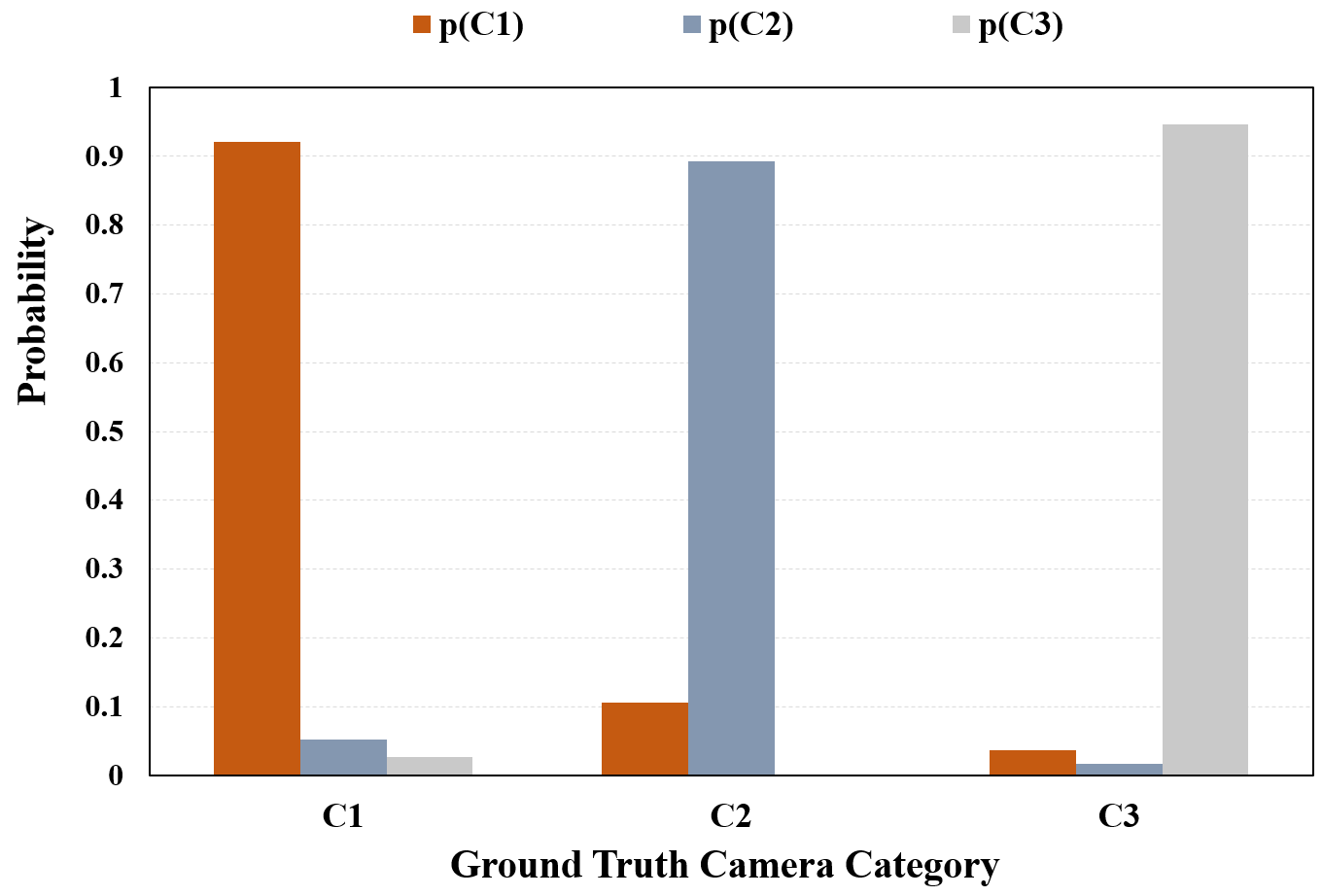}
\caption{Visualization of the average probability of 
camera-specific samples in the \(N\) category classification, where \(N\) is the number of camera category in the training set. The samples for training are a portion of CASIA-FASD dataset and the rest of samples are used for testing. Again, the probability \(p(\cdot)\) corresponds to the probability that the testing sample is classified into the corresponding category. }
\label{fig:pinta}
\end{center}
\end{figure}

\begin{figure*}[ht]
\begin{center}
\includegraphics[width=1.0\textwidth]{ 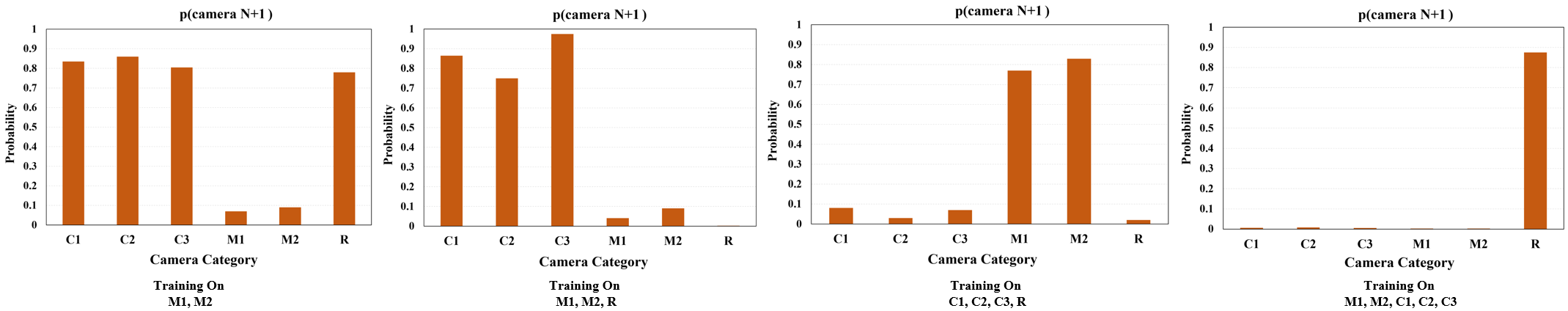}
\caption{Visualization of the average probability of classifying the testing samples to the  category \(N+1\).  The datasets for training are provided under each sub-figure and all samples in six datasets are used for testing.}
\label{fig:np1}
\end{center}
\end{figure*}

\begin{figure}[ht]
\begin{center}
\includegraphics[width=0.75\textwidth]{ 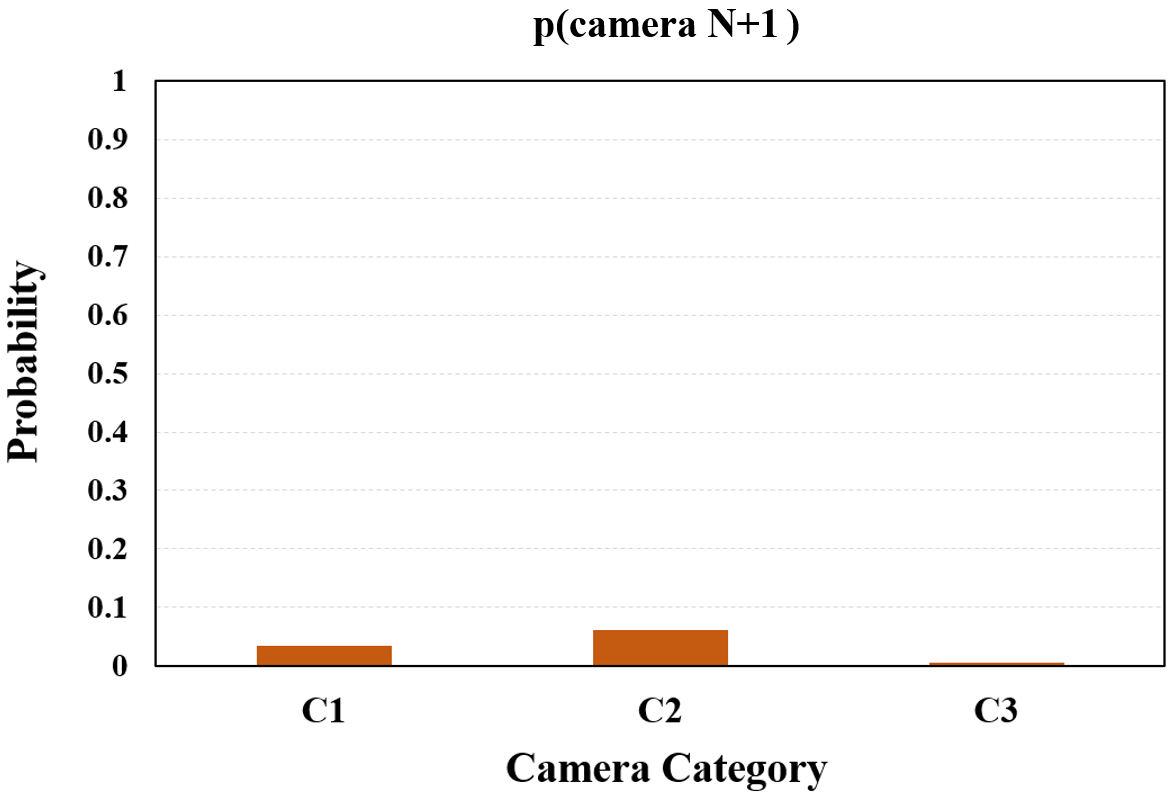}
\caption{ Visualization of the average probability of classifying the testing samples to the  category \(N+1\). The samples for training are a portion of CASIA-FASD dataset and the rest of samples are used for testing.}
\label{fig:n1inta}
\end{center}
\end{figure}

\textit{\textbf{Unknown Camera feature refining.}}
After the camera classification is finished for ``\(N+1\)” classification, an attention based unknown camera feature refining is conducted when the testing camera is classified into the \((N+1)-th\) category (\textit{a.k.a} unknown camera).

Firstly, we use the normalized probability acquired in above step to examine the affiliation of the extracted camera features with respect to each one of \((N+1)\) classes as follows:
\begin{equation}\label{} 
c^t = \underset{c}{argmax} \{p(y^t=c)|c \in \{1,2,...,N+1\}\},
\end{equation}
where $c^t $ is the category that $x^{t}$ is finally classified by comparing all probabilities and selecting the maximum one. When $c^t$ equals to $N+1$, the image is regarded to be captured by an unknown camera. In this scenario, we aim to further improve the detection accuracy based on the philosophy of re-weighting the branches and refining the features. 

1) In the weighted fusion, we increase the weight of results acquired by the invariant branch form 0.7 to 1.0, due to the higher generalization ability of the Invariant branch  compared with Augmentation branch.

2) The extracted camera feature is refined due to unseen noise possibly introduced.  For simplification, we use $X^{cam}$ and $X^{mix}$ to represent $M_{cam}$ and $M_{mix}$ in Fig.~\ref{fig:3} respectively. The $X^{mix}$ is a combination of  camera invariant spoofing feature and camera feature. When the $X^{cam}$ is aligned with the camera information in  $X^{mix}$, the camera invariant spoofing feature can be acquired. To reduce the noise injected in $X^{cam}$, an attention based strategy is proposed. More specifically, as shown in Fig.~\ref{fig:att},  we adopt a self-attention module to measure the spatial attention map \(H\) of $X^{cam}$ as follows:
\begin{equation}\label{att} 
H_{u, v}=\frac{e^{\mathrm{G}_{u, v}}}{\sum_{i=1}^{h w} e^{\mathrm{G}_{u, v}}}.
\end{equation}
where \(h,w\) indicate the spatial size of  \(H\) and \(u,v \) are the indexes of row and column respectively. The matrix \(G\) is given by,
\begin{equation}
G=\left(X^{c a m}\right)^{\top} \times X^{c a m},
\end{equation}
where the ``$\times$" operation represents the matrix multiply.  In particular, for each index in the spatial dimension, we calculate the normalized weights of itself and the other indices to synthesize the refined camera feature as follows:
\begin{equation}
X^{\mathrm{att}}= X^{\text {cam}} \times H.
\end{equation}
Based on the refined camera feature $X^{\mathrm{att}}$,  the feature subtraction is performed, and the binary classification result is finally obtained.   {To verify the  refining strategy, we compare the performance of 
our model with and without (W/O) unknown camera feature refining on different cross-dataset settings. The results are shown in Table~\ref{tab:refc}. From the table, we can observe that the unknown camera feature refining strategy can be benefit for the performance improvement on the most of cross-dataset settings, which demonstrates that the features can be further enhanced to improve the effectiveness of our method.}

\begin{figure}[ht]
\begin{center}
\includegraphics[width=1\textwidth]{ 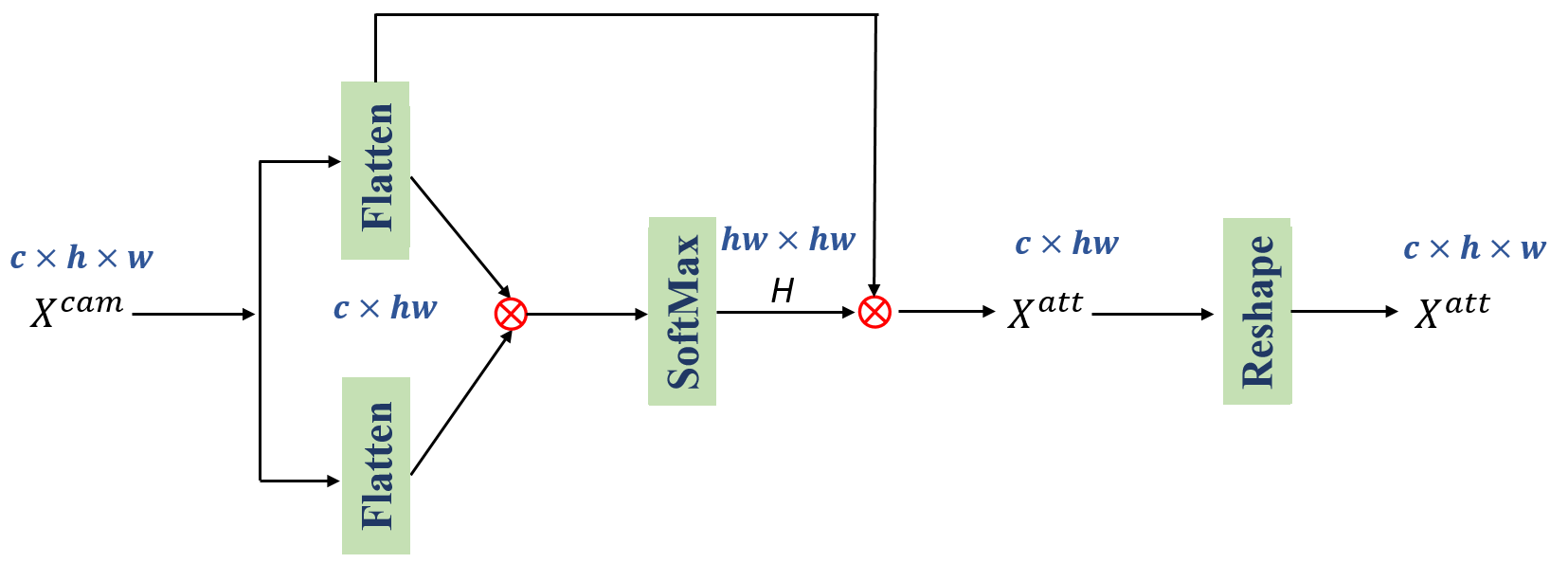}
\caption{Illustration of the attention based unknown camera feature refining. Herein, \textit{h}, \textit{w} and \textit{c} represent the height, width, and number of channels respectively.}
\label{fig:att}
\end{center}
\end{figure}

\begin{table*}
\begin{floatrow}
\scalebox{0.9}{
  \centering
  \caption{Performance  (HTER \%) comparison between our models with and without (W/O) unknown camera feature refining on different cross-dataset settings.}
\begin{tabular}{c|ccccccc}
\hline
\textbf{Method} & \textbf{C $\rightarrow$ R} & \textbf{M $\rightarrow$ R} & \textbf{M $\rightarrow$ C} & \textbf{C $\rightarrow$ M} & \textbf{C\&M $\rightarrow$ R} & \textbf{C\&R $\rightarrow$ M} & \textbf{M\&R $\rightarrow$ C} \bigstrut[b]\\
\hline
\hline
Ours (W/O Unknown Camera Refining)& 17.6  & 21.7  & 27.3  & 17.5  & 21.3  & \textbf{14.8}  & 32.3 \bigstrut\\
Ours (Unknown Camera Refining) & \textbf{17.0}  & \textbf{21.0}  & \textbf{27.1}  & \textbf{13.8}  & \textbf{19.9}  & 15.3  & \textbf{30.7} \bigstrut\\
\hline
\end{tabular}%
 \label{tab:refc}%
}
\end{floatrow}
\end{table*}%

}
\subsection{Investigations in Re-compression Scenarios} 
The compression artifacts will be injected when the video clips are represented in compact format,  leading to the negative effect on spoofing detection. Moreover, with the rapid development of cloud computing, the face anti-spoofing algorithms are also expected  to  be  deployed  on  the  cloud  platforms  and  the  compression/re-compression  could  significantly  reduce  the  bandwidth  when  transmitting  the  video  stream  to  the  cloud.  As such, the evaluation of the spoofing detection performance with videos been compressed/re-compressed is important and essential.  

Herein, we conduct performance comparisons in the scenarios that the testing video data are further recompressed with different codecs based on HEVC standard (x265) \cite{sullivan2012overview, x265} and H.264/AVC standard (x264) \cite{ wiegand2003overview, x264}.  Three databases are adopted in the performance comparisons,  including  CASIA-FASD,  MSU-MFSD  and  REPLAY-ATTACK.  Besides  different  compression standards explored,  multiple quantization parameters (QP) are also selected,  ranging from high to low bit rate (QP=17 to QP=42).  The performance comparisons are shown in Tables~\ref{tab:x265} and ~\ref{tab:x264}.     {In  particular,  it  is  apparent  that  our proposed  method  achieves  better detection performance for different QP values.  However, it is generally acknowledged that the compression artifacts may influence the performance of the detection accuracy. For example, the detection accuracy decreases with the increase of QP, as very important clues that are useful for spoofing detection could be distorted during compression.   {Herein, we also show the unknown camera feature refining results in Table~\ref{tab:x265} and Table~\ref{tab:x264} where the rows are denoted ``Ours (Unknown Camera Refining)". We can observe that the performance in terms of HTER\% results is  further improved. For the high QP scenario ($QP \geq 32 $), the refining strategy is still effective which reveals the robustness of our scheme.}}

\begin{table*}
\begin{floatrow}
\scalebox{0.85}{
  \centering
  \caption{Performance comparisons of different face spoofing methods when the videos are compressed with HEVC standard (x265).}
\begin{tabular}{c|l|ccccccc}
\toprule
\textbf{Configuration} & \textbf{Methods} & \multicolumn{1}{l}{\textbf{Original}} & \multicolumn{1}{l}{\textbf{QP17}} & \multicolumn{1}{l}{\textbf{QP22}} & \multicolumn{1}{l}{\textbf{QP27}} & \multicolumn{1}{l}{\textbf{QP32}} & \multicolumn{1}{l}{\textbf{QP37}} & \multicolumn{1}{l}{\textbf{QP42}} \\
\midrule
\midrule
\multirow{5}[2]{*}{\textbf{CASIA Intra}} & \textbf{LBP+SVM \cite{boulkenafet2016face2}} & 7.5   & 10.8  & 16.2  & 20.6  & 16.9  & 22.2  & 28.8  \\
      & \textbf{Resnet18 \cite{he2016deep}} & 5.2   & 7.6   & 8.3   & 8.1   & 8.7   & 9.5   & 10.9  \\
      & \textbf{DeepPixel \cite{george2019deep} } & 2.6   & 3.2   & 3.0   & 3.5   & 3.7   & 5.2   & 4.4  \\
      & \textbf{Auxiliary (depth only) \cite{liu2018learning}} & 1.3   & 2.0   & \textbf{1.8 } & 2.9   & 3.9   & 7.3   & 9.4  \\
      & \textbf{Ours} & \textbf{ 0.9 } & \textbf{ 1.3 } & 1.9   & \textbf{ 1.6 } & \textbf{ 2.8 } & \textbf{ 3.4 } & \textbf{ 4.2 } \\

\midrule
\multirow{5}[2]{*}{\textbf{C $\rightarrow$ R}} & \textbf{LBP+SVM \cite{boulkenafet2016face2}} & 28.9  & 28.6  & 32.6  & 40.3  & 44.5  & 42.8  & 46.8  \\
      & \textbf{Resnet18 \cite{he2016deep}} & 47.0  & 47.6  & 47.8  & 48.5  & 49.6  & 49.1  & 50.0  \\
      & \textbf{DeepPixel \cite{george2019deep}} & 41.5  & 38.2  & 37.6  & 36.2  & 36.2  & 36.9  & 36.7  \\
      & \textbf{Auxiliary (depth only) \cite{liu2018learning}} & 28.1  & 26.2  & 25.6  & 28.4  & 32.5  & 37.8  & 43.6  \\
      & \textbf{Ours} & 17.6  & 18.7  & 18.7  & 18.9  & 20.6  & 22.8  & 20.2  \\
      
      &   {\textbf{Ours (Unknown Camera Refining)}} &   {\textbf{ 17.0}}  &   {\textbf{ 17.2 }} &   {\textbf{ 16.8 }} &   {\textbf{ 16.4 }} &   {\textbf{ 18.7 }} &   {\textbf{ 21.9 }}&   {\textbf{ 19.9}} \\
\midrule
\multirow{5}[2]{*}{\textbf{C $\rightarrow$ M}} & \textbf{LBP+SVM \cite{boulkenafet2016face2}} & 48.1  & 48.5  & 48.8  & 49.3  & 48.4  & 45.7  & 47.4  \\
      & \textbf{Resnet18 \cite{he2016deep}} & 36.4  & 37.2  & 36.0  & 36.0  & 34.3  & 35.8  & 36.5  \\
      & \textbf{DeepPixel \cite{george2019deep}} & 40.3  & 40.1  & 39.0  & 37.7  & 39.4  & 39.6  & 40.1  \\
      & \textbf{Auxiliary (depth only) \cite{liu2018learning}} & 36.1  & 30.3  & 29.1  & 28.5  & 27.6  & 28.1  & 24.8  \\
      & \textbf{Ours} & 17.5  & 18.0  & 17.5  & 17.8  & 17.6 & 17.3  & 17.2  \\
      
      &   {\textbf{Ours (Unknown Camera Refining)}} &   {\textbf{13.8 }} &   {\textbf{ 13.67 }} &   {\textbf{ 12.6 }} &   {\textbf{ 12.6 }} &   {\textbf{ 12.5 }} &   {\textbf{ 12.2 }} &   {\textbf{ 11.8 }} \\
\midrule
\multirow{5}[2]{*}{\textbf{M $\rightarrow$ R}} & \textbf{LBP+SVM \cite{boulkenafet2016face2}} & 46.4  & 47.8  & 46.5  & 42.1  & 41.0  & 41.8  & 45.1  \\
      & \textbf{Resnet18 \cite{he2016deep}} & 47.9  & 45.1  & 44.8  & 45.9  & 46.0  & 46.4  & 44.1  \\
      & \textbf{DeepPixel \cite{george2019deep}} & 21.9  & 22.7  & 22.7  & 22.6  & \textbf{23.8 } & \textbf{26.4 } & \textbf{28.0 } \\
      & \textbf{Auxiliary(depth only) \cite{liu2018learning}} & 26.7  & 28.0  & 27.6  & 26.5  & 28.1  & 30.9  & 35.2  \\
      & \textbf{Ours} & 21.7  & 22.1  & 23.2  & 19.8  & 25.7  & 30.3  & 36.0  \\
      
      &   {\textbf{Ours (Unknown Camera Refining)}} &   {\textbf{ 21.0 }} &   {\textbf{ 19.9 } }&   {\textbf{ 19.5 }} &   { \textbf{19.2}  }&   { 24.3 } &   { 28.9 } &   { 30.1 } \\
\midrule
\multirow{5}[1]{*}{\textbf{M $\rightarrow$ C}} & \textbf{LBP+SVM \cite{boulkenafet2016face2}} & 49.8  & 50.0  & 49.2  & 49.7  & 48.9  & 50.0  & 48.3  \\
      & \textbf{Resnet18 \cite{he2016deep}} & 45.8  & 48.4  & 48.3  & 48.6  & 48.7  & 47.9  & 49.0  \\
      & \textbf{DeepPixel \cite{george2019deep}} & 36.1  & 34.6  & 35.2  & 34.5  & 35.4  & 36.3  & 36.9  \\
      & \textbf{Auxiliary (depth only) \cite{liu2018learning}} & 44.5  & 38.5  & 37.9  & 38.2  & 39.3  & 41.3  & 42.8  \\
      & \textbf{Ours} & 27.3 & 29.8  & 30.4  & 30.7  & 29.1  & 28.5  & 31.3  \\
      
      &   {\textbf{Ours (Unknown Camera Refining)}} &   {\textbf{ 27.1 }} &   {\textbf{ 28.7 } }&   { \textbf{ 30.1 } }&   {\textbf{ 30.2 }} &   {\textbf{ 28.7 }} &   {\textbf{ 28.3 } }&   {\textbf{ 29.6 }} \\
\midrule
\end{tabular}%

  \label{tab:x265}%
}
\end{floatrow}
\end{table*}%

\begin{table*}
\begin{floatrow}
\scalebox{0.85}{
  \centering
  \caption{Performance comparisons of different face spoofing methods when the videos are compressed with H.264/AVC standard (x264).}
\begin{tabular}{c|l|ccccccc}
\toprule
\textbf{Configuration} & \textbf{Methods} & \multicolumn{1}{l}{\textbf{Original}} & \multicolumn{1}{l}{\textbf{QP17}} & \multicolumn{1}{l}{\textbf{QP22}} & \multicolumn{1}{l}{\textbf{QP27}} & \multicolumn{1}{l}{\textbf{QP32}} & \multicolumn{1}{l}{\textbf{QP37}} & \multicolumn{1}{l}{\textbf{QP42}} \\
\midrule
\midrule
\multirow{5}[1]{*}{\textbf{CASIA Intra}} & \textbf{LBP+SVM \cite{boulkenafet2016face2}} & 7.5   & 8.7   & 32.8  & 25.3  & 32.0  & 28.0  & 32.9  \\
      & \textbf{Resnet18 \cite{he2016deep}} & 5.2   & 7.4   & 7.1   & 7.9   & 9.0   & 8.7   & 11.8  \\
      & \textbf{DeepPixel \cite{george2019deep}} & 2.6   & 3.4   & 3.5   & 3.3   & 3.7   & 5.2   & 7.3  \\
      & \textbf{Auxiliary (depth only) \cite{liu2018learning}} & 1.3   & 1.8   & 2.5   & 2.3   & 4.2   & 6.1   & 11.3  \\
      & \textbf{Ours} & \textbf{0.9 } & \textbf{1.2 } & \textbf{1.7 } & \textbf{1.8 } & \textbf{1.9 } & \textbf{2.3 } & \textbf{5.9 } \\
\midrule
\multirow{5}[2]{*}{\textbf{C $\rightarrow$ R}} & \textbf{LBP+SVM \cite{boulkenafet2016face2}} & 28.9  & 35.4  & 36.3  & 41.7  & 47.4  & 49.5  & 47.3  \\
      & \textbf{Resnet18 \cite{he2016deep} } & 47.0  & 47.5  & 47.4  & 48.8  & 49.1  & 50.6  & 51.3  \\
      & \textbf{DeepPixel \cite{george2019deep}} & 41.5  & 39.3  & 38.8  & 38.6  & 38.1  & 39.8  & 39.7  \\
      & \textbf{Auxiliary (depth only) \cite{liu2018learning}} & 28.1  & 28.2  & 27.0  & 28.9  & 34.6  & 39.3  & 42.8  \\
      & \textbf{Ours}  & 17.6  & 21.9  & 18.5  & 18.2  & 21.6  & 20.8  & 22.1  \\
      
      &   {\textbf{Ours (Unknown Camera Refining)}} &   {\textbf{ 17.0 }} &   {\textbf{ 17.5 }} &   {\textbf{ 17.5 }} &   {\textbf{ 17.1 }} &   {\textbf{ 20.2 }} &   {\textbf{ 20.0 }} &   {\textbf{ 21.5 }} \\
\midrule
\multirow{5}[2]{*}{\textbf{C $\rightarrow$ M}} & \textbf{LBP+SVM \cite{boulkenafet2016face2}} & 48.1  & 46.6  & 47.8  & 49.5  & 49.8  & 48.9  & 48.1  \\
      & \textbf{Resnet18 \cite{he2016deep}} & 36.4  & 36.8  & 36.1  & 35.0  & 35.0  & 35.5  & 34.2  \\
      & \textbf{DeepPixel\cite{george2019deep} } & 40.3  & 41.2  & 40.2  & 39.7  & 41.0  & 42.2  & 42.8  \\
      & \textbf{Auxiliary (depth only) \cite{liu2018learning}} & 36.1  & 30.2  & 28.9  & 28.2  & 28.4  & 27.3  & 27.1  \\
      & \textbf{Ours} & 17.5  & 16.8  & 17.0  & 16.8  & 17.2  & 17.7  & 16.8  \\
      
      &   {\textbf{Ours (Unknown Camera Refining)}} &  {\textbf{13.8 }} &   {\textbf{ 12.9 }} &   {\textbf{ 12.7 }} &   {\textbf{ 12.2 }} &   {\textbf{ 11.8 }} &   {\textbf{ 12.9 }} &   {\textbf{ 11.9 }} \\
\midrule
\multirow{5}[2]{*}{\textbf{M $\rightarrow$ R}} & \textbf{LBP+SVM \cite{boulkenafet2016face2}} & 46.4  & 45.7  & 45.0  & 43.1  & 41.9  & 42.2  & 45.9  \\
      & \textbf{Resnet18 \cite{he2016deep}} & 47.9  & 45.1  & 45.1  & 45.9  & 47.2  & 46.1  & 44.9  \\
      & \textbf{DeepPixel \cite{george2019deep}} & 21.9  & 23.1  & 23.2  & 23.8  & \textbf{ 24.6}  & \textbf{ 28.1}  & \textbf{ 29.9 }\\
      & \textbf{Auxiliary (depth only) \cite{liu2018learning}} & 26.7  & 27.5  & 27.2  & 28.1  & 31.4  & 36.2  & 40.8  \\
      & \textbf{Ours} & 21.7  & 21.9  & 24.5  & 21.9 & 27.0  & 34.0  & 36.7  \\
      
      &   {\textbf{Ours (Unknown Camera Refining)} }  &   {\textbf{ 21.0 } } &   {\textbf{ 21.0 } } &   {\textbf{ 19.9 } } & \textbf{  { 19.5  }} &   { 26.3  } &   { 28.7  } &   { 31.3   }\\
      
\midrule
\multirow{5}[1]{*}{\textbf{M $\rightarrow$ C}} & \textbf{LBP+SVM \cite{boulkenafet2016face2}} & 49.8  & 49.8  & 48.8  & 49.2  & 48.8  & 49.2  & 49.9  \\
      & \textbf{Resnet18 \cite{he2016deep}} & 45.8  & 47.8  & 48.3  & 49.8  & 48.6  & 48.8  & 50.0  \\
      & \textbf{DeepPixel \cite{george2019deep} } & 36.1  & 34.5  & 35.1  & 34.9  & 34.7  & 36.9  & 39.6  \\
      & \textbf{Auxiliary (depth only) \cite{liu2018learning}} & 44.5  & 38.7  & 38.3  & 39.0  & 39.9  & 41.8  & 44.4  \\
      & Ours & 27.3  & 30.3  & 30.2  & 29.9  & 30.4  & 30.7  & 30.9 \\
      
      &   {\textbf{Ours (Unknown Camera Refining)}} &   {\textbf{27.1 }} &   {\textbf{29.8 }} &   {\textbf{29.1 }} &   {\textbf{29.2 }} &   {\textbf{29.8 }} &   {\textbf{29.6 }} &   {\textbf{30.1 }} \\
\midrule
\end{tabular}%

  \label{tab:x264}%
}
\end{floatrow}
\end{table*}%

\section{Conclusions and Future Work}

We aim to address one critical and practical challenge faced by face anti-spoofing that the diverse camera types are prone to cause a large domain gap when the training and testing data are from different cameras.  
 In this paper, we present a novel camera invariant face anti-spoofing model, targeting to improve the generalization capability of face anti-spoofing in real applications. Our model is able to eliminate the influence of cameras in feature extraction based on feature domain decomposition, and practically obtain promising spoofing detection performance based on the combination of camera invariant and discrimination augmented feature extraction. Extensive experiments based on intra-database and cross-database settings verify that the proposed scheme achieves superior performance and exhibits strong generalization capability in face spoofing detection.
 
Recent years have witnessed a surge of growth in terms of camera types along with different mobile devices. 
It is envisioned that the proposed method could be naturally incorporated to support the applications in these devices. Based on the elaborate design, more accurate face anti-spoofing is achieved without specifically collecting and labeling training data from the given camera, as evidenced by our experimental results. The proposed method is also extensible when more environments, attack types and compression levels are involved in the face spoofing detection. 
As such, how the proposed model could be further extended to a more unified spoofing detection model, is an interesting direction yet to be explored.
It is also anticipated that the design philosophy could facilitate many other applications in addition to face spoofing detection. One concrete example is the low-level computer vision task such as image quality assessment, as a model that exhibits strong robustness to different cameras and acquisition environments is highly desired. Another example is the high-level visual understanding at the cloud side, as a universal model that works across different cameras can be deployed in the cloud for efficient recognition and understanding.

\section{Acknowledgement}
The authors would like to thank the anonymous reviewers
for their valuable comments which greatly helped improve this
paper.


\bibliographystyle{IEEEtran}
\bibliography{main}

\begin{thebibliography}{10}
\providecommand{\url}[1]{#1}
\csname url@samestyle\endcsname
\providecommand{\newblock}{\relax}
\providecommand{\bibinfo}[2]{#2}
\providecommand{\BIBentrySTDinterwordspacing}{\spaceskip=0pt\relax}
\providecommand{\BIBentryALTinterwordstretchfactor}{4}
\providecommand{\BIBentryALTinterwordspacing}{\spaceskip=\fontdimen2\font plus
\BIBentryALTinterwordstretchfactor\fontdimen3\font minus
  \fontdimen4\font\relax}
\providecommand{\BIBforeignlanguage}[2]{{%
\expandafter\ifx\csname l@#1\endcsname\relax
\typeout{** WARNING: IEEEtran.bst: No hyphenation pattern has been}%
\typeout{** loaded for the language `#1'. Using the pattern for}%
\typeout{** the default language instead.}%
\else
\language=\csname l@#1\endcsname
\fi
#2}}
\providecommand{\BIBdecl}{\relax}
\BIBdecl

\bibitem{chingovska2012effectiveness}
I.~Chingovska, A.~Anjos, and S.~Marcel, ``On the effectiveness of local binary
  patterns in face anti-spoofing,'' in \emph{2012 BIOSIG-proceedings of the
  international conference of biometrics special interest group
  (BIOSIG)}.\hskip 1em plus 0.5em minus 0.4em\relax IEEE, 2012, pp. 1--7.

\bibitem{boulkenafet2016face}
Z.~Boulkenafet, J.~Komulainen, and A.~Hadid, ``Face antispoofing using
  speeded-up robust features and fisher vector encoding,'' \emph{IEEE Signal
  Processing Letters}, vol.~24, no.~2, pp. 141--145, 2016.

\bibitem{komulainen2013context}
J.~Komulainen, A.~Hadid, and M.~Pietik{\"a}inen, ``Context based face
  anti-spoofing,'' in \emph{2013 IEEE Sixth International Conference on
  Biometrics: Theory, Applications and Systems (BTAS)}.\hskip 1em plus 0.5em
  minus 0.4em\relax IEEE, 2013, pp. 1--8.

\bibitem{patel2016secure}
K.~Patel, H.~Han, and A.~K. Jain, ``Secure face unlock: Spoof detection on
  smartphones,'' \emph{IEEE transactions on information forensics and
  security}, vol.~11, no.~10, pp. 2268--2283, 2016.

\bibitem{yang2014learn}
J.~Yang, Z.~Lei, and S.~Z. Li, ``Learn convolutional neural network for face
  anti-spoofing,'' \emph{arXiv preprint arXiv:1408.5601}, 2014.

\bibitem{patel2016cross}
K.~Patel, H.~Han, and A.~K. Jain, ``Cross-database face antispoofing with
  robust feature representation,'' in \emph{Chinese Conference on Biometric
  Recognition}.\hskip 1em plus 0.5em minus 0.4em\relax Springer, 2016, pp.
  611--619.

\bibitem{li2016original}
L.~Li, X.~Feng, Z.~Boulkenafet, Z.~Xia, M.~Li, and A.~Hadid, ``An original face
  anti-spoofing approach using partial convolutional neural network,'' in
  \emph{2016 Sixth International Conference on Image Processing Theory, Tools
  and Applications (IPTA)}.\hskip 1em plus 0.5em minus 0.4em\relax IEEE, 2016,
  pp. 1--6.

\bibitem{feng2016integration}
L.~Feng, L.-M. Po, Y.~Li, X.~Xu, F.~Yuan, T.~C.-H. Cheung, and K.-W. Cheung,
  ``Integration of image quality and motion cues for face anti-spoofing: A
  neural network approach,'' \emph{Journal of Visual Communication and Image
  Representation}, vol.~38, pp. 451--460, 2016.

\bibitem{liu2018learning}
Y.~Liu, A.~Jourabloo, and X.~Liu, ``Learning deep models for face
  anti-spoofing: Binary or auxiliary supervision,'' in \emph{Proceedings of the
  IEEE Conference on Computer Vision and Pattern Recognition}, 2018, pp.
  389--398.

\bibitem{liu20163d}
S.~Liu, P.~C. Yuen, S.~Zhang, and G.~Zhao, ``3d mask face anti-spoofing with
  remote photoplethysmography,'' in \emph{European Conference on Computer
  Vision}.\hskip 1em plus 0.5em minus 0.4em\relax Springer, 2016, pp. 85--100.

\bibitem{li2018unsupervised}
H.~Li, W.~Li, H.~Cao, S.~Wang, F.~Huang, and A.~C. Kot, ``Unsupervised domain
  adaptation for face anti-spoofing,'' \emph{IEEE Transactions on Information
  Forensics and Security}, vol.~13, no.~7, pp. 1794--1809, 2018.

\bibitem{li2018learning}
H.~Li, P.~He, S.~Wang, A.~Rocha, X.~Jiang, and A.~C. Kot, ``Learning
  generalized deep feature representation for face anti-spoofing,'' \emph{IEEE
  Transactions on Information Forensics and Security}, vol.~13, no.~10, pp.
  2639--2652, 2018.

\bibitem{tzeng2017adversarial}
E.~Tzeng, J.~Hoffman, K.~Saenko, and T.~Darrell, ``Adversarial discriminative
  domain adaptation,'' in \emph{Proceedings of the IEEE Conference on Computer
  Vision and Pattern Recognition}, 2017, pp. 7167--7176.

\bibitem{zhang2012face}
Z.~Zhang, J.~Yan, S.~Liu, Z.~Lei, D.~Yi, and S.~Z. Li, ``A face antispoofing
  database with diverse attacks,'' in \emph{2012 5th IAPR international
  conference on Biometrics (ICB)}.\hskip 1em plus 0.5em minus 0.4em\relax IEEE,
  2012, pp. 26--31.

\bibitem{he2016deep}
K.~He, X.~Zhang, S.~Ren, and J.~Sun, ``Deep residual learning for image
  recognition,'' in \emph{Proceedings of the IEEE conference on computer vision
  and pattern recognition}, 2016, pp. 770--778.

\bibitem{chen2012bayesian}
D.~Chen, X.~Cao, L.~Wang, F.~Wen, and J.~Sun, ``Bayesian face revisited: A
  joint formulation,'' in \emph{European conference on computer vision}.\hskip
  1em plus 0.5em minus 0.4em\relax Springer, 2012, pp. 566--579.

\bibitem{jourabloo2018face}
A.~Jourabloo, Y.~Liu, and X.~Liu, ``Face de-spoofing: Anti-spoofing via noise
  modeling,'' in \emph{Proceedings of the European Conference on Computer
  Vision (ECCV)}, 2018, pp. 290--306.

\bibitem{maaten2008visualizing}
L.~v.~d. Maaten and G.~Hinton, ``Visualizing data using t-sne,'' \emph{Journal
  of machine learning research}, vol.~9, no. Nov, pp. 2579--2605, 2008.

\bibitem{wen2015face}
D.~Wen, H.~Han, and A.~K. Jain, ``Face spoof detection with image distortion
  analysis,'' \emph{IEEE Transactions on Information Forensics and Security},
  vol.~10, no.~4, pp. 746--761, 2015.

\bibitem{evans2019handbook}
N.~Evans, \emph{Handbook of Biometric Anti-Spoofing: Presentation Attack
  Detection}.\hskip 1em plus 0.5em minus 0.4em\relax Springer, 2019.

\bibitem{li2004live}
J.~Li, Y.~Wang, T.~Tan, and A.~K. Jain, ``Live face detection based on the
  analysis of fourier spectra,'' in \emph{Biometric Technology for Human
  Identification}, vol. 5404.\hskip 1em plus 0.5em minus 0.4em\relax
  International Society for Optics and Photonics, 2004, pp. 296--303.

\bibitem{de2014face}
T.~de~Freitas~Pereira, J.~Komulainen, A.~Anjos, J.~M. De~Martino, A.~Hadid,
  M.~Pietik{\"a}inen, and S.~Marcel, ``Face liveness detection using dynamic
  texture,'' \emph{EURASIP Journal on Image and Video Processing}, vol. 2014,
  no.~1, p.~2, 2014.

\bibitem{chan2017face}
P.~P. Chan, W.~Liu, D.~Chen, D.~S. Yeung, F.~Zhang, X.~Wang, and C.-C. Hsu,
  ``Face liveness detection using a flash against 2d spoofing attack,''
  \emph{IEEE Transactions on Information Forensics and Security}, vol.~13,
  no.~2, pp. 521--534, 2017.

\bibitem{galbally2014face}
J.~Galbally and S.~Marcel, ``Face anti-spoofing based on general image quality
  assessment,'' in \emph{2014 22nd International Conference on Pattern
  Recognition}.\hskip 1em plus 0.5em minus 0.4em\relax IEEE, 2014, pp.
  1173--1178.

\bibitem{galbally2013image}
J.~Galbally, S.~Marcel, and J.~Fierrez, ``Image quality assessment for fake
  biometric detection: Application to iris, fingerprint, and face
  recognition,'' \emph{IEEE transactions on image processing}, vol.~23, no.~2,
  pp. 710--724, 2013.

\bibitem{boulkenafet2018generalization}
Z.~Boulkenafet, J.~Komulainen, and A.~Hadid, ``On the generalization of color
  texture-based face anti-spoofing,'' \emph{Image and Vision Computing},
  vol.~77, pp. 1--9, 2018.

\bibitem{pan2008liveness}
G.~Pan, Z.~Wu, and L.~Sun, ``Liveness detection for face recognition,'' in
  \emph{Recent advances in face recognition}.\hskip 1em plus 0.5em minus
  0.4em\relax IntechOpen, 2008.

\bibitem{tirunagari2015detection}
S.~Tirunagari, N.~Poh, D.~Windridge, A.~Iorliam, N.~Suki, and A.~T. Ho,
  ``Detection of face spoofing using visual dynamics,'' \emph{IEEE transactions
  on information forensics and security}, vol.~10, no.~4, pp. 762--777, 2015.

\bibitem{bharadwaj2013computationally}
S.~Bharadwaj, T.~I. Dhamecha, M.~Vatsa, and R.~Singh, ``Computationally
  efficient face spoofing detection with motion magnification,'' in
  \emph{Proceedings of the IEEE conference on computer vision and pattern
  recognition workshops}, 2013, pp. 105--110.

\bibitem{siddiqui2016face}
T.~A. Siddiqui, S.~Bharadwaj, T.~I. Dhamecha, A.~Agarwal, M.~Vatsa, R.~Singh,
  and N.~Ratha, ``Face anti-spoofing with multifeature videolet aggregation,''
  in \emph{2016 23rd International Conference on Pattern Recognition
  (ICPR)}.\hskip 1em plus 0.5em minus 0.4em\relax IEEE, 2016, pp. 1035--1040.

\bibitem{chingovska20132nd}
I.~Chingovska, J.~Yang, Z.~Lei, D.~Yi, S.~Z. Li, O.~Kahm, C.~Glaser, N.~Damer,
  A.~Kuijper, A.~Nouak \emph{et~al.}, ``The 2nd competition on counter measures
  to 2d face spoofing attacks,'' in \emph{2013 International Conference on
  Biometrics (ICB)}.\hskip 1em plus 0.5em minus 0.4em\relax IEEE, 2013, pp.
  1--6.

\bibitem{wang2013face}
T.~Wang, J.~Yang, Z.~Lei, S.~Liao, and S.~Z. Li, ``Face liveness detection
  using 3d structure recovered from a single camera,'' in \emph{2013
  international conference on biometrics (ICB)}.\hskip 1em plus 0.5em minus
  0.4em\relax IEEE, 2013, pp. 1--6.

\bibitem{wang2017robust}
Y.~Wang, F.~Nian, T.~Li, Z.~Meng, and K.~Wang, ``Robust face anti-spoofing with
  depth information,'' \emph{Journal of Visual Communication and Image
  Representation}, vol.~49, pp. 332--337, 2017.

\bibitem{zhang2011face}
Z.~Zhang, D.~Yi, Z.~Lei, and S.~Z. Li, ``Face liveness detection by learning
  multispectral reflectance distributions,'' in \emph{Face and Gesture
  2011}.\hskip 1em plus 0.5em minus 0.4em\relax IEEE, 2011, pp. 436--441.

\bibitem{manjani2017detecting}
I.~Manjani, S.~Tariyal, M.~Vatsa, R.~Singh, and A.~Majumdar, ``Detecting
  silicone mask-based presentation attack via deep dictionary learning,''
  \emph{IEEE Transactions on Information Forensics and Security}, vol.~12,
  no.~7, pp. 1713--1723, 2017.

\bibitem{pinto2018counteracting}
A.~Pinto, H.~Pedrini, M.~Krumdick, B.~Becker, A.~Czajka, K.~W. Bowyer, and
  A.~Rocha, ``Counteracting presentation attacks in face, fingerprint, and iris
  recognition,'' \emph{Deep Learning in Biometrics}, vol. 245, 2018.

\bibitem{parkhi2015deep}
O.~M. Parkhi, A.~Vedaldi, and A.~Zisserman, ``Deep face recognition,''
  \emph{British Machine Vision Association}, 2015.

\bibitem{atoum2017face}
Y.~Atoum, Y.~Liu, A.~Jourabloo, and X.~Liu, ``Face anti-spoofing using patch
  and depth-based cnns,'' in \emph{2017 IEEE International Joint Conference on
  Biometrics (IJCB)}.\hskip 1em plus 0.5em minus 0.4em\relax IEEE, 2017, pp.
  319--328.

\bibitem{wang2019improving}
G.~Wang, H.~Han, S.~Shan, and X.~Chen, ``Improving cross-database face
  presentation attack detection via adversarial domain adaptation,'' in
  \emph{International Conference on Biometrics (ICB)}, 2019.

\bibitem{tu2019learning}
X.~Tu, J.~Zhao, M.~Xie, G.~Du, H.~Zhang, J.~Li, Z.~Ma, and J.~Feng, ``Learning
  generalizable and identity-discriminative representations for face
  anti-spoofing,'' \emph{arXiv preprint arXiv:1901.05602}, 2019.

\bibitem{LiSource}
C.~T. Li, ``Source camera identification using enhanced sensor pattern noise,''
  \emph{IEEE Transactions on Information Forensics and Security}, vol.~5,
  no.~2, pp. 280--287.

\bibitem{Sutcu2007Improvements}
Y.~Sutcu, S.~Bayram, H.~T. Sencar, and N.~Memon, ``Improvements on sensor noise
  based source camera identification,'' in \emph{2007 IEEE International
  Conference on Multimedia and Expo}.\hskip 1em plus 0.5em minus 0.4em\relax
  IEEE, 2007, pp. 24--27.

\bibitem{chen2007digital}
M.~Chen, J.~Fridrich, and M.~Goljan, ``Digital imaging sensor identification
  (further study),'' in \emph{Security, steganography, and watermarking of
  multimedia contents IX}, vol. 6505.\hskip 1em plus 0.5em minus 0.4em\relax
  International Society for Optics and Photonics, 2007, p. 65050P.

\bibitem{zagoruyko2015learning}
S.~Zagoruyko and N.~Komodakis, ``Learning to compare image patches via
  convolutional neural networks,'' in \emph{Proceedings of the IEEE conference
  on computer vision and pattern recognition}, 2015, pp. 4353--4361.

\bibitem{zhang2016joint}
K.~Zhang, Z.~Zhang, Z.~Li, and Y.~Qiao, ``Joint face detection and alignment
  using multitask cascaded convolutional networks,'' \emph{IEEE Signal
  Processing Letters}, vol.~23, no.~10, pp. 1499--1503, 2016.

\bibitem{fridrich2012rich}
J.~Fridrich and J.~Kodovsky, ``Rich models for steganalysis of digital
  images,'' \emph{IEEE Transactions on Information Forensics and Security},
  vol.~7, no.~3, pp. 868--882, 2012.

\bibitem{zhou2018learning}
P.~Zhou, X.~Han, V.~I. Morariu, and L.~S. Davis, ``Learning rich features for
  image manipulation detection,'' in \emph{Proceedings of the IEEE Conference
  on Computer Vision and Pattern Recognition}, 2018, pp. 1053--1061.

\bibitem{goljan2015cfa}
M.~Goljan and J.~Fridrich, ``Cfa-aware features for steganalysis of color
  images,'' in \emph{Media Watermarking, Security, and Forensics 2015}, vol.
  9409.\hskip 1em plus 0.5em minus 0.4em\relax International Society for Optics
  and Photonics, 2015, p. 94090V.

\bibitem{lin2017focal}
T.-Y. Lin, P.~Goyal, R.~Girshick, K.~He, and P.~Doll{\'a}r, ``Focal loss for
  dense object detection,'' in \emph{Proceedings of the IEEE international
  conference on computer vision}, 2017, pp. 2980--2988.

\bibitem{boulkenafet2017oulu}
Z.~Boulkenafet, J.~Komulainen, L.~Li, X.~Feng, and A.~Hadid, ``Oulu-npu: A
  mobile face presentation attack database with real-world variations,'' in
  \emph{2017 12th IEEE International Conference on Automatic Face \& Gesture
  Recognition (FG 2017)}.\hskip 1em plus 0.5em minus 0.4em\relax IEEE, 2017,
  pp. 612--618.

\bibitem{wu2018group}
Y.~Wu and K.~He, ``Group normalization,'' in \emph{Proceedings of the European
  Conference on Computer Vision (ECCV)}, 2018, pp. 3--19.

\bibitem{BiometricPresentation}
I.~J. S.~. Biometrics, ``Information technology - biometric presentation attack
  detection-part 1: Framework. international organization for
  standardization,'' 2016.

\bibitem{george2019deep}
A.~George and S.~Marcel, ``Deep pixel-wise binary supervision for face
  presentation attack detection,'' in \emph{Proceedings of 2019 International
  Conference on Biometrics (ICB)}.\hskip 1em plus 0.5em minus 0.4em\relax IEEE,
  2019, pp. 1--8.

\bibitem{anjos2011counter}
A.~Anjos and S.~Marcel, ``Counter-measures to photo attacks in face
  recognition: a public database and a baseline,'' in \emph{2011 international
  joint conference on Biometrics (IJCB)}.\hskip 1em plus 0.5em minus
  0.4em\relax IEEE, 2011, pp. 1--7.

\bibitem{boulkenafet2017competition}
Z.~Boulkenafet, J.~Komulainen, Z.~Akhtar, A.~Benlamoudi, D.~Samai, S.~E.
  Bekhouche, A.~Ouafi, F.~Dornaika, A.~Taleb-Ahmed, L.~Qin \emph{et~al.}, ``A
  competition on generalized software-based face presentation attack detection
  in mobile scenarios,'' in \emph{2017 IEEE International Joint Conference on
  Biometrics (IJCB)}.\hskip 1em plus 0.5em minus 0.4em\relax IEEE, 2017, pp.
  688--696.

\bibitem{boulkenafet2016face2}
Z.~Boulkenafet, J.~Komulainen, and A.~Hadid, ``Face spoofing detection using
  colour texture analysis,'' \emph{IEEE Transactions on Information Forensics
  and Security}, vol.~11, no.~8, pp. 1818--1830, 2016.

\bibitem{chen2019attention}
H.~Chen, G.~Hu, Z.~Lei, Y.~Chen, N.~M. Robertson, and S.~Z. Li,
  ``Attention-based two-stream convolutional networks for face spoofing
  detection,'' \emph{IEEE Transactions on Information Forensics and Security},
  vol.~15, pp. 578--593, 2019.

\bibitem{hendrycks2016baseline}
D.~Hendrycks and K.~Gimpel, ``A baseline for detecting misclassified and
  out-of-distribution examples in neural networks,'' \emph{arXiv preprint
  arXiv:1610.02136}, 2016.

\bibitem{yu2019unsupervised}
Q.~Yu and K.~Aizawa, ``Unsupervised out-of-distribution detection by maximum
  classifier discrepancy,'' in \emph{Proceedings of the IEEE International
  Conference on Computer Vision}, 2019, pp. 9518--9526.

\bibitem{mandal2019out}
D.~Mandal, S.~Narayan, S.~K. Dwivedi, V.~Gupta, S.~Ahmed, F.~S. Khan, and
  L.~Shao, ``Out-of-distribution detection for generalized zero-shot action
  recognition,'' in \emph{Proceedings of the IEEE Conference on Computer Vision
  and Pattern Recognition}, 2019, pp. 9985--9993.

\bibitem{sullivan2012overview}
G.~J. Sullivan, J.-R. Ohm, W.-J. Han, and T.~Wiegand, ``Overview of the high
  efficiency video coding (hevc) standard,'' \emph{IEEE Transactions on
  circuits and systems for video technology}, vol.~22, no.~12, pp. 1649--1668,
  2012.

\bibitem{x265}
\url{http://x265.org/}.

\bibitem{wiegand2003overview}
T.~Wiegand, G.~J. Sullivan, G.~Bjontegaard, and A.~Luthra, ``Overview of the h.
  264/avc video coding standard,'' \emph{IEEE Transactions on circuits and
  systems for video technology}, vol.~13, no.~7, pp. 560--576, 2003.

\bibitem{x264}
\url{https://x264.org/en/}.

\end{thebibliography}

\begin{IEEEbiography}[{\includegraphics[width=1in,height=1.25in,clip,keepaspectratio]{ 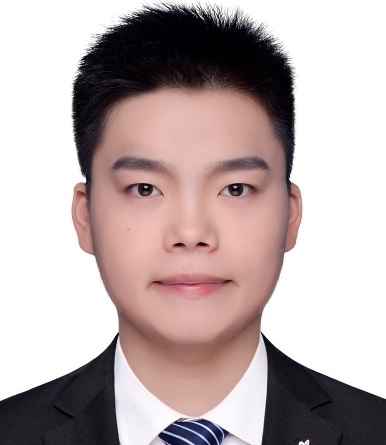}}]{Baoliang Chen} received the B.S. degree in Electronic Information Science and Technology from Hefei University of Technology, Hefei, China, in 2015 and the M.S. degree in Intelligent Information Processing from Xidian University, Xian, China, in 2018. He was a researcher in iFlytek Inc., since 2018. He is currently pursuing the Ph.D. degree in Department of Computer Science of City University of HongKong, HongKong. His current research interests include  image/video quality assessment and information security.
\end{IEEEbiography}

\begin{IEEEbiography}[{\includegraphics[width=1in,height=1.25in,clip,keepaspectratio]{ 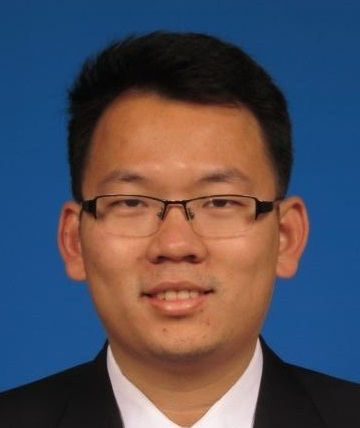}}]
{Wenhan Yang} (Member, IEEE)  	
received the B.S degree and Ph.D. degree (Hons.) in computer science from Peking University, Beijing, China, in 2012 and 2018. He is currently a postdoctoral research fellow with the Department of Computer Science, City University of Hong Kong. Dr. His current research interests include image/video processing/restoration, bad weather restoration, human-machine collaborative coding. He has authored over 100 technical articles in refereed journals and proceedings, and holds 9 granted patents.
He received the IEEE ICME-2020 Best Paper Award, the IFTC 2017 Best Paper Award, and the IEEE CVPR-2018 UG2 Challenge First Runner-up Award.
He was the Candidate of CSIG Best Doctoral Dissertation Award in 2019. He served as the Area Chair of IEEE ICME-2021, and the Organizer of IEEE CVPR-2019/2020/2021 UG2+ Challenge and Workshop.
\end{IEEEbiography}

\begin{IEEEbiography}[{\includegraphics[width=1in,height=1.25in,clip,keepaspectratio]{ 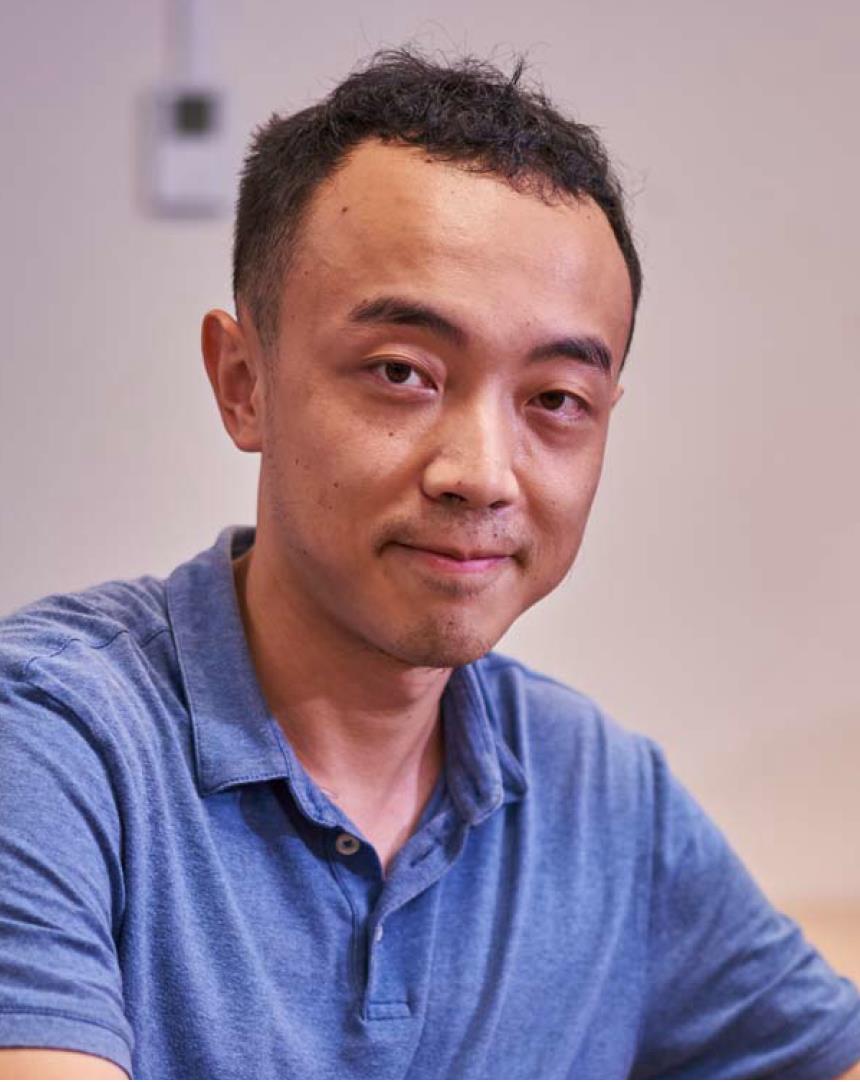}}]{Haoliang Li}  (Member, IEEE) received the B.S. degree from University of Electronic Science and Technology of China in 2013, and the Ph.D. degree from Nanyang Technological University (NTU), Singapore, in 2018. He was a project officer in 2018 and a research fellow from July 2018 to May 2019 in Rapid-Rich Object Search Lab, NTU. He is now a Wallenberg-NTU presidential postdoc fellow in NTU. He received the doctorate innovation award from NTU in 2019. His research interest is information forensics and security, and transfer learning.
\end{IEEEbiography}

\begin{IEEEbiography}[{\includegraphics[width=1in,height=1.25in,clip,keepaspectratio]{ 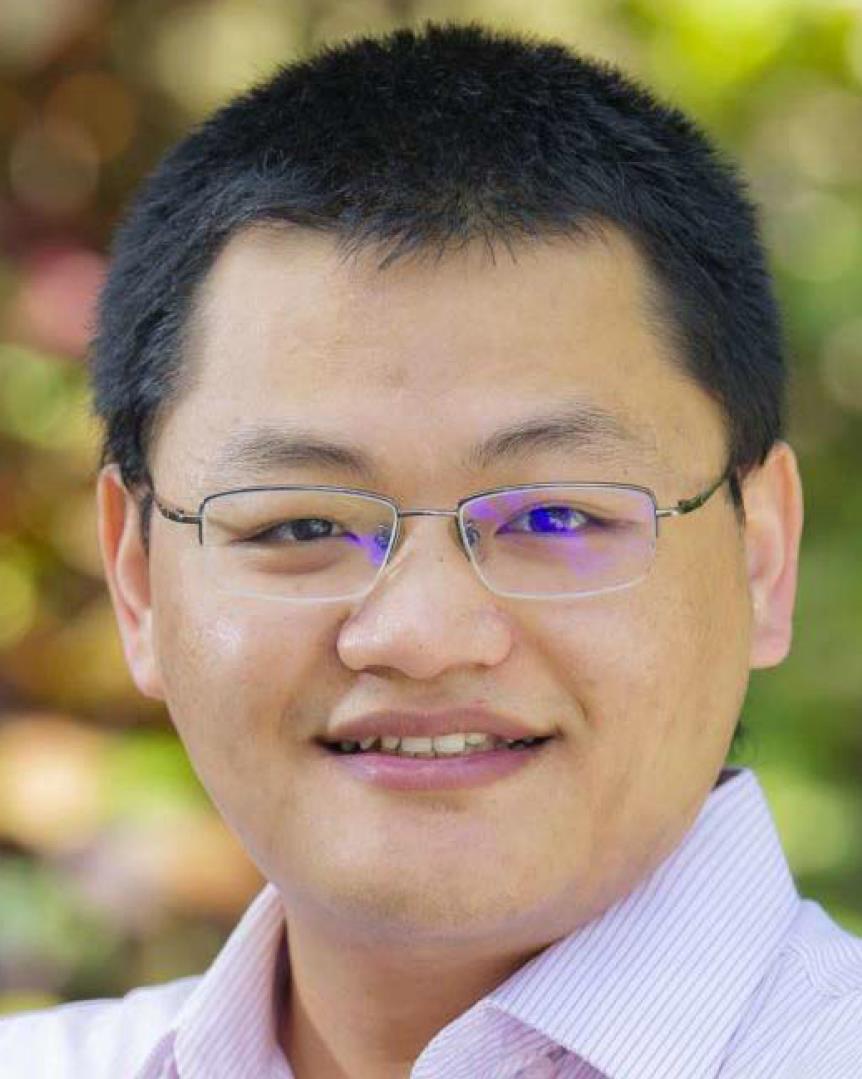}}]{Shiqi Wang}  (Member, IEEE) received the B.S. degree in computer science from the Harbin Institute of Technology in 2008 and the Ph.D. degree in computer application technology from Peking University in 2014. From 2014 to 2016, he was a Post-Doctoral Fellow with the Department of Electrical and Computer Engineering, University of
Waterloo, Waterloo, ON, Canada. From 2016 to
2017, he was a Research Fellow with the Rapid-Rich
Object Search Laboratory, Nanyang Technological
University, Singapore. He is currently an Assistant Professor with the Department of Computer Science, City University of Hong Kong. He has proposed over 40 technical proposals to ISO/MPEG, ITU-T, and AVS standards, and authored/coauthored more than 150 refereed journal articles/conference papers. He received the Best Paper Award from IEEE VCIP 2019, ICME 2019, IEEE Multimedia 2018, and PCM 2017 and is the coauthor of an article that received the Best Student Paper Award in the IEEE ICIP 2018. His research interests include video compression, image/video quality assessment, and image/video search and analysis. 
\end{IEEEbiography}

\begin{IEEEbiography}[{\includegraphics[width=1in,height=1.25in,clip,keepaspectratio]{ 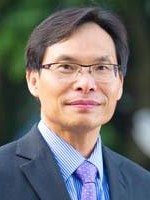}}]{Sam Kwong}(Fellow, IEEE) received the B.S. and M.S. degrees in electrical engineering from State University of New York at Buffalo in 1983, University of Waterloo, Waterloo, ON, Canada, in 1985, and the Ph.D. degree from University of Hagen, Germany, in 1996. From 1985 to 1987, he was a Diagnostic Engineer with Control Data Canada. He joined Bell Northern Research Canada as a Member of Scientific Staff. In 1990, he became a Lecturer at the Department of Electronic Engineering, City University of Hong Kong, where he is currently a Chair Professor of the Department of Computer Science, City University of Hong Kong, Kowloon, China (Hong Kong SAR). His research interests are video and image coding and evolutionary algorithms.
\end{IEEEbiography}

\end{document}